\documentclass{article}

\usepackage{microtype}
\usepackage{graphicx}
\usepackage{subfigure}
\usepackage{booktabs} 

\usepackage{hyperref}

\usepackage[utf8]{inputenc} 
\usepackage[T1]{fontenc}    
\usepackage{hyperref}       
\usepackage{url}            
\usepackage{amsfonts}       
\usepackage{nicefrac}       
\usepackage{xcolor}         

\usepackage{lipsum}
\usepackage{transparent}
\usepackage{multicol}
\usepackage{multirow}
\usepackage{enumitem}
\usepackage{amsmath}
\usepackage{amssymb}
\usepackage{mathtools}
\usepackage{amsthm}
\usepackage[capitalize,noabbrev]{cleveref}

\usepackage[percent]{overpic}
\usepackage{tcolorbox}

\usepackage[accepted]{mlsys2023}

\mlsystitlerunning{RevBiFPN: The Fully Reversible Bidirectional Feature Pyramid Network}

\begin{document}

\twocolumn[
\mlsystitle{
RevBiFPN: The Fully Reversible \\
Bidirectional Feature Pyramid Network
}



\mlsyssetsymbol{equal}{*}

\begin{mlsysauthorlist}
\mlsysauthor{Vitaliy Chiley}{wdwa_cerb}
\mlsysauthor{Vithursan Thangarasa}{cerb}
\mlsysauthor{Abhay Gupta}{cerb}
\mlsysauthor{Anshul Samar}{cerb}
\mlsysauthor{Joel Hestness}{cerb}
\mlsysauthor{Dennis DeCoste}{wdwa_cerb}
\end{mlsysauthorlist}

\mlsysaffiliation{cerb}{Cerebras Systems Inc, California, USA}
\mlsysaffiliation{wdwa_cerb}{Work done while at Cerebras Systems Inc}

\mlsyscorrespondingauthor{Vitaliy Chiley}{vitaliy@mosaicml.com}
\mlsyscorrespondingauthor{Vithursan Thangarasa}{vithu@cerebras.net}
\mlsyscorrespondingauthor{Abhay Gupta}{abhay@cerebras.net}

\mlsyskeywords{Machine Learning, MLSys}

\vskip 0.3in

\begin{abstract}
This work introduces RevSilo, the first reversible bidirectional multi-scale feature fusion module.
Like other reversible methods,
RevSilo eliminates the need to store hidden activations by recomputing them.
However, existing reversible methods do not apply to multi-scale feature fusion and are, therefore, not applicable to a large class of networks.
Bidirectional multi-scale feature fusion promotes local and global coherence and has become a de facto design principle for networks targeting spatially sensitive tasks, e.g., HRNet~\cite{sun2019hrnet_pose} and EfficientDet~\cite{tan2020efficientdet}.
These networks achieve state-of-the-art results across various computer vision tasks when paired with high-resolution inputs. However, training them requires substantial accelerator memory for saving large, multi-resolution activations.
These memory requirements inherently cap the size of neural networks, limiting improvements that come from scale.
Operating across resolution scales, RevSilo alleviates these issues.
Stacking RevSilos, we create RevBiFPN, a fully reversible bidirectional feature pyramid network.
RevBiFPN is competitive with networks such as EfficientNet while using up to 19.8x lesser training memory for image classification.
When fine-tuned on MS COCO, RevBiFPN provides up to a 2.5\% boost in AP over HRNet using fewer MACs and a 2.4x reduction in training-time memory.
\end{abstract}
]

\printAffiliationsAndNotice{}

\section{Introduction}
\label{sec:intro}

\begin{figure}
    \centering
    \vskip -7pt
    \begin{overpic}[width=\linewidth]
        {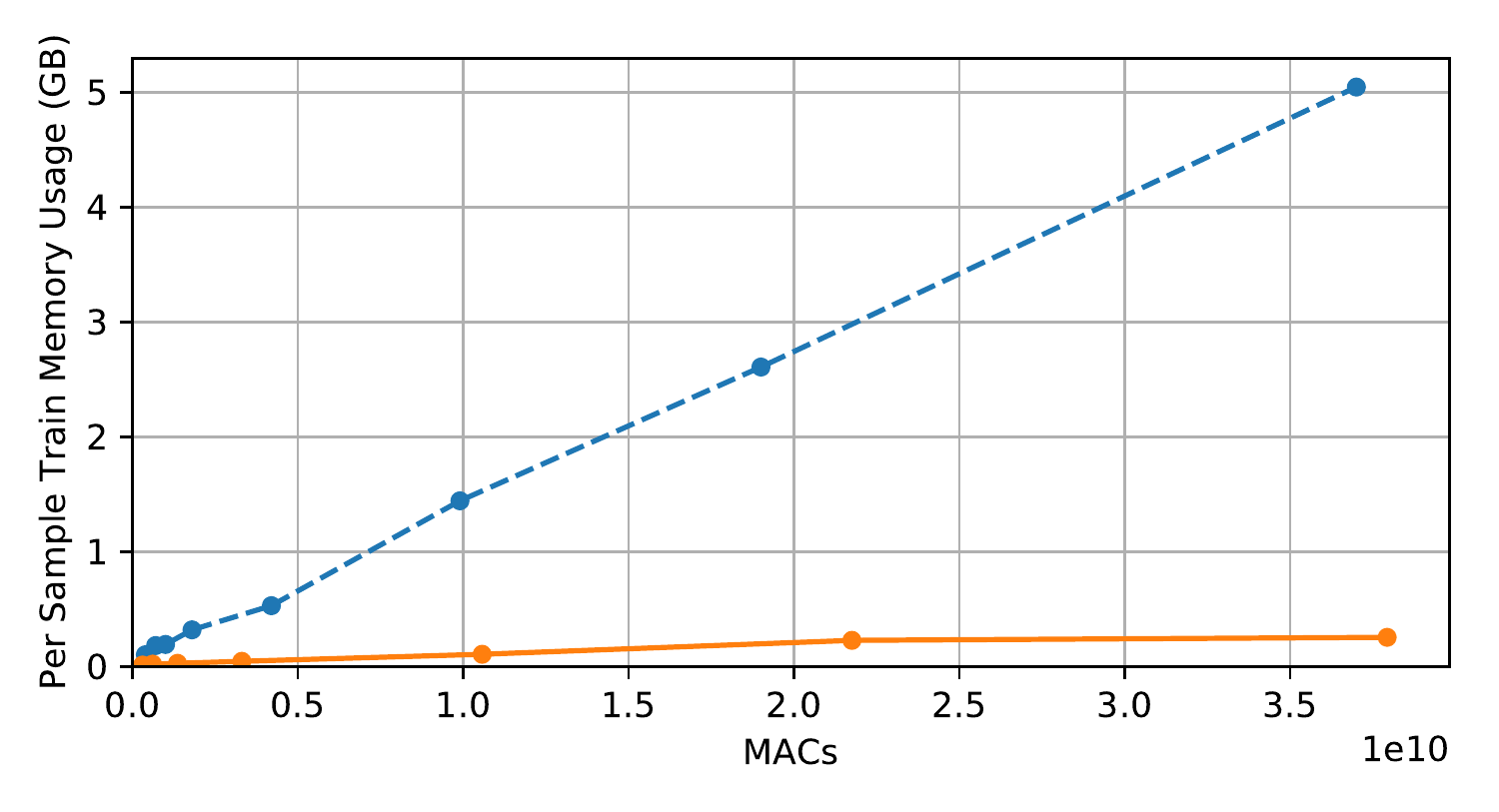}
        \put(8.2,30.6){
            \tcbox[size=minimal,colback=white,colframe=white]{
            \scalebox{0.45}{
            \centering
            \small
            \sc
          
            \begin{tabular}{lrrr}
                \toprule
                Model                 &   Top1 & MACs &           Mem \\ \midrule 
                \textbf{RevBiFPN-S4} & 83.0\% &  11B & \textbf{0.11} \\         
                EfficientNet-B5       & 83.6\% &  10B &         1.44  \\ \midrule 
                \textbf{RevBiFPN-S5} & 83.7\% &  22B & \textbf{0.23} \\          
                EfficientNet-B6       & 84.0\% &  19B &         2.61  \\ \midrule 
                \textbf{RevBiFPN-S6} & 84.2\% &  38B & \textbf{0.25} \\          
                EfficientNet-B7       & 84.3\% &  37B &         5.05  \\         
                \bottomrule
            \end{tabular}
        }}}
        \put(31,19){ 
            \tcbox[size=minimal,colback=white,colframe=white]{
            \scalebox{0.5}{\textsf{B5}}
        }}
        \put(51,28){ 
            \tcbox[size=minimal,colback=white,colframe=white]{
            \scalebox{0.5}{\textsf{B6}}
        }}
        \put(73,47.5){ 
            \tcbox[size=minimal,colback=white,colframe=white]{
            \scalebox{0.5}{\textsf{EfficientNet-B7}}
        }}
        \put(27,11){ 
            \tcbox[size=minimal,colback=white,colframe=white]{
            \scalebox{0.5}{\textsf{S4}}
        }}
        \put(52.5,12){ 
            \tcbox[size=minimal,colback=white,colframe=white]{
            \scalebox{0.5}{\textsf{S5}}
        }}
        \put(78,12.5){ 
            \tcbox[size=minimal,colback=white,colframe=white]{
            \scalebox{0.5}{\textsf{RevBiFPN-S6}}
        }}
    \end{overpic}
    \vskip -10pt
    \caption{
        \textbf{MACs vs. Measured Memory Usage for ImageNet Training}:
        RevBiFPN significantly outperforms EfficientNet at all scales.
        In particular, RevBiFPN-S6 achieves comparable accuracy (84.2\%) to EfficientNet-B7 on ImageNet while using comparable MACs (38.1B) and 19.8x lesser training memory per sample.
        Details in \cref{tab:mem,tab:i1kacc_ext}.
    }
    \label{fig:memvscompute}
    \vskip -20pt
\end{figure}

State-of-the-art (SOTA) computer vision (CV) networks have large memory requirements that complicate training and limit scalability.
\citet{tan2019efficientnet} and \citet{dollar2021fast} show how compound scaling, i.e., scaling input resolution, network width, and depth, results in efficient networks across a wide range of parameters and MAC (multiply-accumulate) counts.
Even when resources are optimally allocated, scaling networks produce large feature maps. Thus, training requires a large amount of accelerator memory (\cref{fig:memvscompute}).
While low-resolution intermediate representations work well for classification tasks \cite{leCun1998gradient, krizhevsky2012imagenet, simonyan2014very, tan2019efficientnet}, dense prediction tasks, such as detection and segmentation, require the construction of spatially informative, high-resolution feature maps which further exacerbates memory issues.

\begin{figure*}
    \centering
    \includegraphics[width=\textwidth]{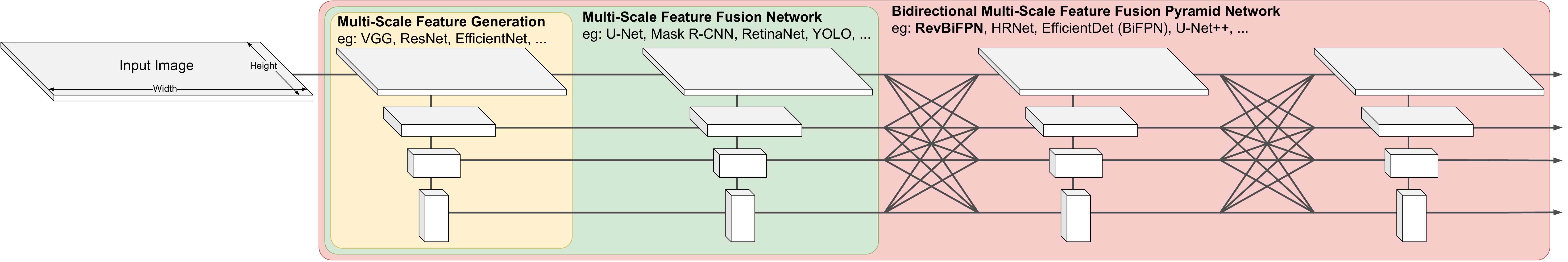}
    \vskip -2pt
    \caption{
    \textbf{Connectivity of multi-scale networks}:
    Features are depicted as boxes, and the lines represent the possible connectivity of networks processing features at multiple scales.
    Networks like VGG~\cite{simonyan2014very}, ResNet~\cite{he2016deep}, and EfficientNet~\cite{tan2019efficientnet} can generate multi-scale features (yellow box).
    These features are often fused by networks such as U-Net~\cite{ronneberger2015unet}, Mask R-CNN~\cite{he2017mask}, or YOLO~\cite{redmon2018yolov3} for completing spatially sensitive tasks (green box). 
    Low-resolution features communicate global information, while high-resolution features capture detailed local features such as texture and object boundaries.
    By iteratively mixing these features, bidirectional multi-scale feature fusion networks such as HRNet~\cite{sun2019hrnet_pose}, EfficientDet~\cite{tan2020efficientdet}, and UNet++~\cite{zhou2018unet} promote local and global coherence, boosting performance (red box). See \cref{sec:cnnectivity_ex} for more details.}
    \label{fig:bifpn}
    \vskip -10pt
\end{figure*}

U-Net~\cite{ronneberger2015unet}, used for segmentation, was one of the first multi-scale feature fusion networks.
Initially, detection networks would perform multi-scale inference by processing every scale of an image pyramid independently. But, they soon adopted multi-scale feature fusion to directly produce a feature pyramid, i.e., multi-scale features \cite{lin2017feature, lin2017focal, redmon2018yolov3}.
Bidirectional multi-scale feature fusion networks iteratively merge information between high and low-resolution feature maps, producing robust \cite{hendrycks2019benchmarking} scale-invariant models.
These models promote local and global coherence by iteratively aligning the semantic representations of fine-grained and high-level features.
As a result, these networks are often the backbone of SOTA computer vision systems \cite{liu2018path, cai2018cascade, sun2019hrnet_pose, tan2020efficientdet}. But, the memory requirements of backpropagating through multi-scale feature fusion complicate training and limit scalability.
Training CV networks push the memory bounds of modern hardware, with hardware memory setting a hard limit on how far researchers scale these networks, enforcing an upper bound on network performance.

Hidden activations are needed to compute the gradient of the loss with respect to a neural network's parameters.
Traditionally, the activations computed during the forward pass are cached for use in the backward pass.
While this method of neural network training has worked well in the past, the growth of neural networks has outpaced increases in accelerator memory.
Motivated by flow structures \cite{dinh2016density,kingma2018glow}, \citet{gomez2017reversible} recognized that if a network is designed using a series of invertible operations, the activations can be recomputed during the backward pass.
Using this paradigm, reversible networks perform ``backpropagation without storing activations'' \cite{gomez2017reversible}, reducing their activation memory complexity with respect to depth from linear to constant.
Although reversible structures have been successfully used in image classification \cite{gomez2017reversible,jacobsen2018irevnet} and language modeling \cite{kitaev2020reformer,mackay2018reversible}, they have yet to be used where they are needed most:
in multi-scale feature fusion networks to produce high-resolution feature maps.

\subsection{Contributions}
\label{sec:intro:contrib}
To address the memory challenges of training models for spatially sensitive tasks, this work introduces the RevSilo and the network built with it, RevBiFPN.
The main contributions of this work are:

\begin{enumerate}[noitemsep, leftmargin=15pt]
    \item The RevSilo (\cref{fig:c_silo}), the first bidirectional multi-scale feature fusion module that is invertible.
    \item RevBiFPN (\cref{fig:frbifpn}) is the first fully reversible bidirectional multi-scale feature fusion pyramid network. It is built using the RevSilo and uses a fraction of the memory compared to the same network without reversible recomputation (\cref{fig:mvd,fig:mvr}).
    \item With a classification head, RevBiFPN is pretrained on ImageNet~\cite{deng2009imagenet} to accuracies competitive with networks designed specifically for classification (\cref{fig:memvscompute} and \cref{sec:exp:cls}).
    \item To our knowledge, this work is the first to fine-tune a reversible backbone on downstream CV tasks. With the appropriate heads, RevBiFPN is competitive with similar networks on detection and segmentation tasks while using a fraction of the accelerator memory for training (\cref{sec:exp:coco}).
\end{enumerate}

The reference implementation and model checkpoints for ImageNet are available at \url{https://github.com/CerebrasResearch/RevBiFPN}.

\section{Background}
\label{sec:bgnd}

\begin{table*}
    \caption{
        Memory and computational complexity of different memory-saving methods with respect to the depth of the network (D).
        When training in layer pipeline mode \cite{ptrowski1993performanceAO, kosson2021pipelined}, activation complexity is quadratic with respect to depth. While gradient checkpointing decreases activation memory complexity from $O(D^2)$ to $O(D^{1.5})$ \cite{yang2021pipemare}, reversible recomputation decreases it to $O(D)$. Both methods have a $O(D)$ overhead in the backward pass to re-materialize or recompute the activations.
    }
    \vskip 10pt
    \centering
    \small
    \sc
    \begin{tabular}{lcccc}
        \toprule
                                      &  \multicolumn{2}{c}{Memory}            &               \multicolumn{2}{c}{Compute} \\ 
                                      & Layer Sequential &  Pipelined Parallel &  Forward Pass & Backward Pass                     \\ \midrule
        SGD Baseline                  &           $O(D)$ &           $O(D^2)$  &        $O(D)$ & $O(2D)$ \\
        with Checkpointing            &    $O(\sqrt{D})$ & $O(D^\frac{3}{2})$  & $O(2D)$ & $O(2D)$ \\
        with Reversble Recomputation  &           $O(1)$ &             $O(D)$  & $O(2D)$ & $O(2D)$ \\
        \bottomrule
    \end{tabular}
    \label{tab:mem_comp_complex}
\end{table*}

Systems using low-resolution features were often applied to image pyramids for detection \cite{girshick2015fast, ren2015faster, redmon2016you, redmon2017yolo9000}.
\citet{lin2017feature} augment a pretrained classification network with a low-resolution to a high-resolution decoder to perform multi-scale feature fusion similar to the U-Net design.
Rather than a single high-resolution feature map, the network outputs features from multiple spatial resolutions to create a feature pyramid.
The success of the Feature Pyramid Network (FPN) motivated similar methodologies to be used throughout the computer vision community \cite{redmon2018yolov3, bochkovskiy2020yolov4, he2017mask, lin2017focal, goyal2021nondeep}.
Bidirectional multi-scale feature fusion pyramid networks (BiFPNs) further improve performance by iteratively applying multi-scale feature fusion modules \cite{tan2020efficientdet, ghiasi2019nasfpn, liu2018path, cai2018cascade, chen2018cascaded}.
This allows local information from high-resolution feature maps to be repeatedly fused with global information from low-resolution feature maps (\cref{fig:bifpn}).

FPNs are often created using feature fusion modules to augment existing classification networks that are not designed for feature fusion.
However,~\citet{zhou2015interlinked},
\citet{jacobsen2017multiscale},
\citet{ke2017multigrid},
\citet{huang2018multiscale},
\citet{sun2019hrnet_pose},
\citet{sun2019high_wp},
\citet{wang2020hrnet},
\citet{cheng2020higherhrnet},
\citet{fan2021multiscalevit},
and
\citet{li2021improved}
advocate for treating bidirectional multi-scale feature fusion as a first-class design principle in computer vision networks and show the effectiveness of this approach for classification, detection, and segmentation.
Bidirectional multi-scale feature fusion networks reduce the semantic gap between consecutive feature maps \cite{zhou2018unet}.
By outputting a feature pyramid, these networks are also scale invariant.
This improves performance, but their memory requirements complicate training and limit scalability.

Achieving SOTA results frequently requires bidirectional multi-scale feature fusion pyramid networks, or BiFPN style networks, to process mega-pixel images.
This can result in a single sample's activations consuming all accelerator memory \cite{tao2020hierarchical}.
Distributed training setups can accelerate these workloads but impose other limitations.
For instance, using small batch sizes precludes using Batch Normalization~\cite{ioffe2015batch}, requiring different normalization methods \cite{wu2018group, chiley2019online, rao2020batch, labatie2021proxy}.
Alternatively, researchers can adopt model parallel approaches to scaling models, but this often results in hardware utilization or network optimization issues \cite{huang2019gpipe, chen2018efficient, narayanan2019pipedream, kosson2021pipelined}.
Another way to alleviate accelerator memory usage is to offload activations to host \cite{rajbhandari2021zero}.
However, for bandwidth-constrained systems, this results in poor FLOP utilization.
When performing operations with low arithmetic intensity, such as non-linearities or depthwise convolutions \cite{lu2021optimizing, qin2018diagonalwise}, limited device bandwidth memory already results in poor FLOP utilization.
Offloading activations to host uses bandwidth which is even further constrained, exacerbating the issue.

Alternatively, \citet{volin1985automatic}, \citet{griewank2000algorithm}, \citet{zweig2000exact}, \citet{lewis2003debugging}, \citet{dauvergne2006data}, \citet{griewank2008evaluating}, \citet{gruslys2016memory}, \citet{chen2016training}, \citet{jain2016checkmate}, and \citet{fenghuang2021} 
propose gradient (or reverse) checkpointing where a subset of activations are recomputed instead of being stored.
Network activations are needed to compute parameter gradients.
Storing them produces an activation memory complexity that is linear in network depth.
Checkpointing can reduce this complexity from $O(D)$ to $O(\sqrt{D})$ \cite{chen2016training}.

Reversible models \cite{gomez2017reversible, mackay2018reversible, brugger2019partially, pendse2020memory, yamazaki2021invertible, sander2021momentum, chun2020momentum, kitaev2020reformer, nestler2021homebrewnlp} save memory by recomputing activations instead of storing them.
This decreases the activation memory complexity from linear to constant.
Reversible recomputation enables SOTA research without needing hardware with the latest memory capacity, which prolongs the useful life of existing hardware.
As a result, less e-waste is produced, but it comes at the cost of recomputing activations, contributing to the carbon footprint of training reversible models.
\cref{tab:mem_comp_complex} shows the theoretical compute and memory complexity of training neural networks with different memory-saving techniques.
\cref{sec:exp:cls} shows the effect this has in practice.

It should be noted that reversible networks relying on Reversible Residual Block (RevBlock) \cite{gomez2017reversible} are not fully reversible.
RevBlock cannot operate across different dimensionalities. Therefore RevNet \cite{gomez2017reversible} and other networks built using RevBlocks, must cache activations in computational blocks that change tensor shape.
Fully reversible models have the added benefit of being used for generation with Normalizing Flows \cite{dinh2014nice, germain2015made, dinh2016density, kingma2016improved, papamakarios2017masked, kingma2018glow, huang2018neural, jacobsen2018irevnet, keller2021self} but are often not as efficient.
For instance, the injective variant of i-RevNet~\cite{jacobsen2018irevnet} is a fully reversible variant of RevNet but requires a 7x increase in size to match RevNet's performance.
Other approaches to reversible recomputation impose architectural limits \cite{bai2019deep}, limit optimization \cite{behrmann2019invertible, thangarasa2019reversible}, or are computationally expensive \cite{behrmann2019invertible}.
While any reversible model or method could be used for saving activation memory, none were previously applicable to bidirectional multi-scale feature fusion.

As existing reversible structures keep tensor dimensionality constant, they cannot be directly applied to multi-scale networks such as EfficientDet.
One approach to producing high-resolution feature maps would be to apply the reversible residual block~\cite{gomez2017reversible} to an entire subnetwork, such as each hourglass of the Stacked Hourglass Network~\cite{newell2016stacked}.
While feasible, the entire subnetwork of activations would still need to be stored, limiting memory savings (\cref{appx:rev_sh:mem}).
The specific case of the hourglass design also produces high MAC count networks (\cref{appx:rev_sh:FLOP}) and does not provide bidirectional multi-scale feature fusion with a feature pyramid output.

\section{Reversible Residual Silo}
\label{sec:revsilo}

\begin{figure}[h]
    \centering
    \includegraphics[width=.8\linewidth]{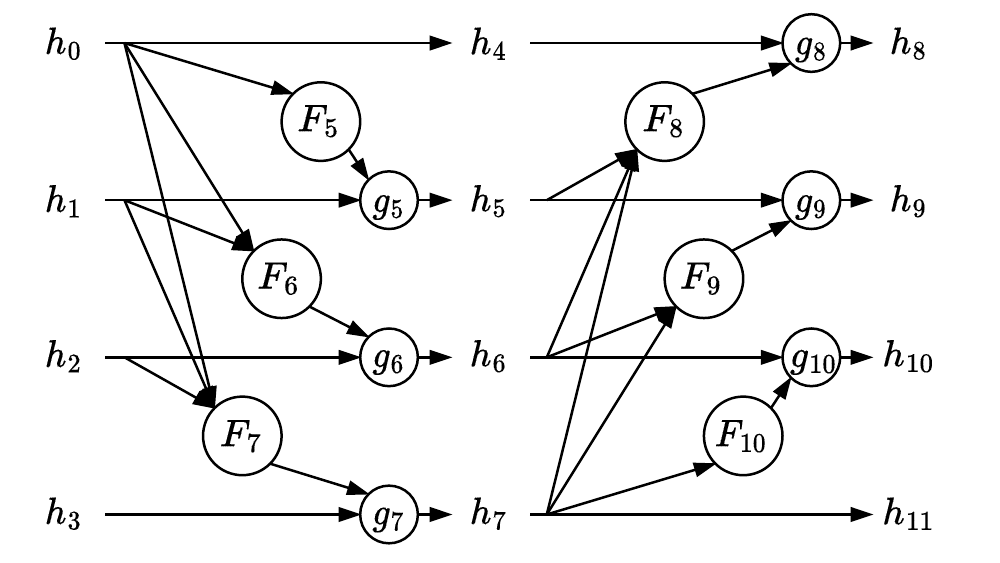}
    \caption{
        \textbf{An $N=4$ RevSilo}.
        $F_j$ can be any transformation; $g_j$ can be any potentially parameterized, invertible transformation.
    }
    \label{fig:c_silo}
    \vskip -5pt
\end{figure}

The Reversible Residual Silo, or RevSilo, generalizes both affine coupling \cite{dinh2014nice} and the reversible residual block \cite{gomez2017reversible} to create an invertible module for bidirectional multi-scale feature fusion.
\cref{fig:c_silo} shows the two halves of the RevSilo with $N = 4$ spatial resolutions.

The left half communicates information down the feature pyramid, and the right half sends information up the feature pyramid.
$g_j$ can be any potentially parameterized, invertible transformation.
In this work, $g_j$ is element-wise addition, and therefore its inverse, $g_j^{-1}$, is element-wise subtraction for all $j$ (\cref{appx:a_revsilo}).
If $h_i$ and $h_j$ are on the same row, i.e., $i \mathbin{\%} N == j \mathbin{\%} N$, the RevSilo's residual structure requires that the shape of $h_i$ equals the shape of $h_j$.
Otherwise, $F_j$ should transform the shape of its inputs to match the shape of $h_j$.
Besides this shape constraint, $F_j$ can be any transformation.
The RevSilo construct remains invertible even if some inputs are $0$.
Setting $h_3$ to $0$ can, for example, be used to expand an $N=3$ feature pyramid into an $N=4$ feature pyramid.
The equations for the $N = 4$ RevSilo are:
\begin{align}
    h_{ 4} &= h_{0} \label{eqn:h0} \\
    h_{ 5} &= g_{5}\left(h_{1}, F_{ 5}(h_0)\right) \\
    h_{ 6} &= g_{6}\left(h_{2}, F_{ 6}(h_1, h_0)\right) \\
    h_{ 7} &= g_{7}\left(h_{3}, F_{ 7}(h_2, h_1, h_0)\right) \label{eqn:h7}
\end{align}
followed by:
\begin{align}
    h_{ 8} &= g_{8}\left(h_{4}, F_{ 8}(h_7, h_6, h_5)\right) \label{eqn:h8} \\
    h_{ 9} &= g_{9}\left(h_{5}, F_{ 9}(h_7, h_6)\right) \\
    h_{10} &= g_{10}\left(h_{6}, F_{10}(h_7)\right) \\
    h_{11} &= h_{7} \label{eqn:h11} 
\end{align}
The corresponding inverse equations are:
\begin{align}
    h_{7} &= h_{11} \label{eqn:bh7} \\
    h_{6} &= g_{10}^{-1}\left(h_{10}, F_{10}(h_7)\right) \\
    h_{5} &= g_{9}^{-1}\left(h_{9}, F_{ 9}(h_7, h_6)\right) \\
    h_{4} &= g_{8}^{-1}\left(h_{8}, F_{ 8}(h_7, h_6, h_5)\right) \label{eqn:bh4}
\end{align}
followed by:
\begin{align}
    h_{0} &= h_{4} \label{eqn:bh0} \\
    h_{1} &= g_{5}^{-1}\left(h_{5}, F_{ 5}(h_0)\right) \\
    h_{2} &= g_{6}^{-1}\left(h_{6}, F_{ 6}(h_1, h_0)\right) \\
    h_{3} &= g_{7}^{-1}\left(h_{7}, F_{ 7}(h_2, h_1, h_0)\right) \label{eqn:bh3}
\end{align}

For the $N = 4$ RevSilo, \crefrange{eqn:h0}{eqn:h11} are used to compute the forward pass. 
Instead of storing activations, they can be recomputed during the backward pass using \crefrange{eqn:bh7}{eqn:bh3}.
While the inverse equations must be computed in order, the forward equations allow the $N$ hidden tensors of the RevSilo to be computed simultaneously.
This enables more parallelism in the resulting inference network.

It should also be noted that if, for all $i$, $h_i$ is a scalar, $g_j$ is addition, and $F_j$ is the dot product operation for all $j$; then the forward equations can be rewritten as matrix-vector products with unitriangular matrices.
The underlying structure that makes unitriangular matrices invertible \cite{thoma2013solving}, makes all coupling structures \cite{kingma2016improved, germain2015made, papamakarios2017masked, huang2018neural, dinh2014nice, gomez2017reversible} invertible.


\begin{figure*}
    \centering
    \includegraphics[width=\textwidth]{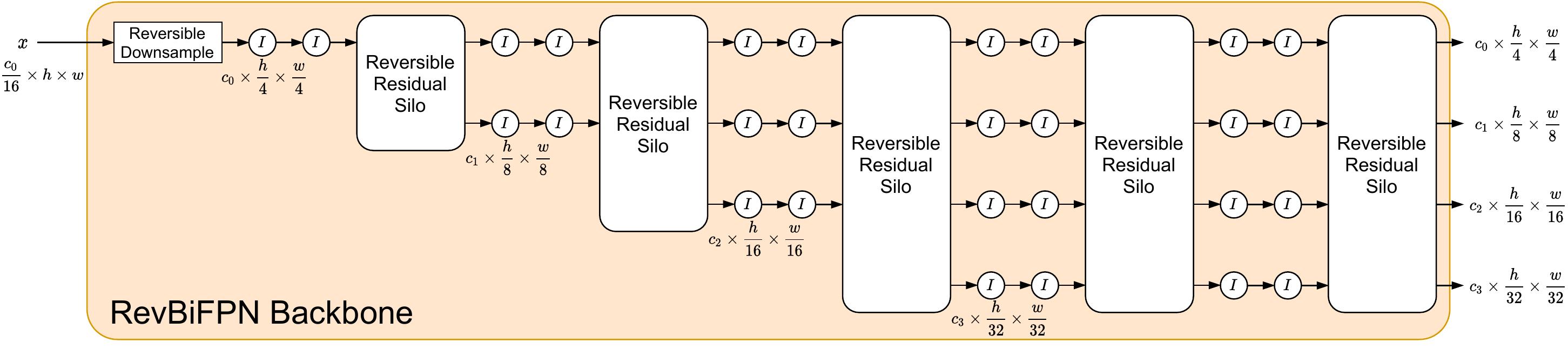}
    \vskip -10pt
    \caption{
        \textbf{A RevBiFPN that creates an $N=4$ feature pyramid}.
        Given the output feature pyramid, all activations can be recomputed going backward through the network. 
        The $I$ components are reversible residual blocks.
        The network builds an $N=4$ multi-scale hidden representation using 3 RevSilos and has an extra depth of $d=2$ RevSilos for further feature fusion.
    }
    \label{fig:frbifpn}
    \vskip -5pt
\end{figure*}

\section{RevBiFPN}
\label{sec:frbifpn}

Reversible Bi-directional Feature Pyramid Network, or RevBiFPN, uses the RevSilo to create a fully reversible backbone that utilizes bidirectional multi-scale feature fusion and produces a feature pyramid output.
Using a reversible multi-scale feature fusion module, RevBiFPN circumvents the issues seen in the RevNet and i-RevNet design (\cref{sec:bgnd}).
The high-level network structure of RevBiFPN is shown in \cref{fig:frbifpn}.
The output feature pyramid can then be used as an input to different task-specific heads (\cref{sec:frbifpn:heads}).

The network uses the invertible SpaceToDepth stem \cite{ridnik2020tresnet, shi2016real, dinh2016density, jacobsen2018irevnet} to initially downsample the input by a factor of $4$ and produce $c = 4^2\times3 = 48$ channels.
The baseline model (RevBiFPN-S0) uses $c_0=48$, $c_1=64$, $c_2=80$, and $c_3=160$ channels in its $N = 4$ spatial resolutions.
As the network size increases, the input image channels are duplicated to ensure the network is fully reversible regardless of network width.
The rest of the network has a structure similar to HRNet~\cite{sun2019hrnet_pose} where transformations in the same spatial resolution use Reversible Residual Blocks~\cite{gomez2017reversible}. 

For simplicity, the network uses the RevSilo variant shown in \cref{fig:a_silo} (\cref{appx:a_revsilo}).
Here, the $F$ operations independently transform and sum each input.
The network isn't designed with a specific hardware target in mind and therefore uses the MBConv block~\cite{howard2017mobilenets}, a building block that efficiently utilizes parameters and MACs (multiply-accumulates).
\footnote{Many research papers often report MACs as FLOPs, which is incorrect. This work uses MAC to mean multiply-accumulate, as FLOP is a single floating point operation.}
The MBConv block is used for both transformations in the reversible residual block and the $F$ transformations of the RevSilo.
Using the MBConv block in network design produces networks with fast inference speed on inference devices \cite{howard2017mobilenets, sandler2018mobilenetv2, howard2019searchingmbconv, tan2019efficientnet,  mehta2021mobilevit}.

Within its RevSilos, RevBiFPN upsamples and downsamples features by factors of 2.
To upsample a feature by a factor of $2^k$, the depthwise convolution of the MBConv block uses a stride of 1 and a kernel size of $3$ or $5$;
this is then followed by bilinear upsampling.
To downsample a feature by a factor of $2^k$, the depthwise convolution of the MBConv block uses a stride of $2^k$ and a kernel size of $2^{k+1} \pm 1$.
As a result, the network uses a diverse set of kernel sizes as suggested by \citet{tan2019mnasnet}.
Network parameters are initialized using Kaiming Initialization \cite{he2015delving}.
Batch Normalization biases are initialized to zero, and weights are initialized to one, except the weights of the last normalization layer, which are initialized to zero to promote stability \cite{kingma2018glow}.

The network uses the MBConv variant with squeeze-excite layers \cite{tan2019efficientnet} and the hard-swish non-linearity \cite{howard2019searchingmbconv}.
The network has larger expansion ratios on the lower resolution streams and uses larger squeeze-excite ratios on the large resolution streams \cite{ridnik2020tresnet}.
With a classification head, the resulting network has a parameter and MAC profile similar to common classification networks.
RevBiFPN-S0 is then scaled (\cref{sec:frbifpn:scaling}) and compared to other network families on the commonly used ImageNet~\cite{deng2009imagenet} benchmark.
The RevBiFPN family of networks is pretrained on ImageNet and fine-tuned with task-specific heads (\cref{sec:frbifpn:heads}) for object detection and instance segmentation on MS COCO~\cite{lin2014microsoft}.


\begin{figure*}
    \begin{minipage}{0.48\linewidth}
    \centering
    \includegraphics[width=\linewidth]{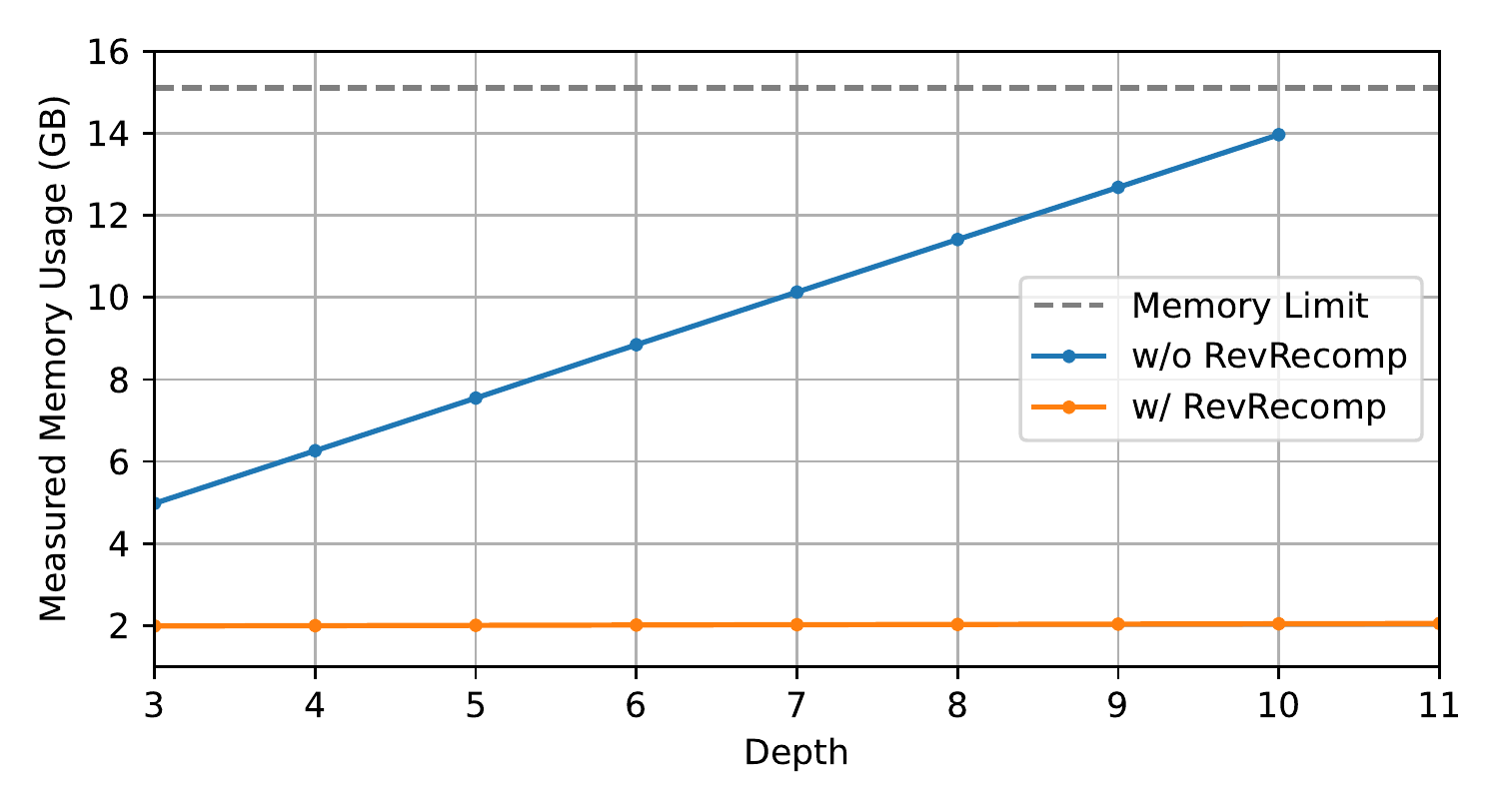}
    \vskip -5pt
    \caption{
        Single GPU memory usage with batch size 64 for training RevBiFPN-S0 on ImageNet with and without reversible recomputation (RevRecomp) as depth is scaled.
        Reversible recomputation decreases the activation memory complexity from linear to constant.
        These memory savings can be reallocated to scaling network width and input resolution to produce RevBiFPN S1-S6.}
    \label{fig:mvd}
    \end{minipage}
    \hfill
    \begin{minipage}{0.48\linewidth}
    \centering
    \includegraphics[width=\linewidth]{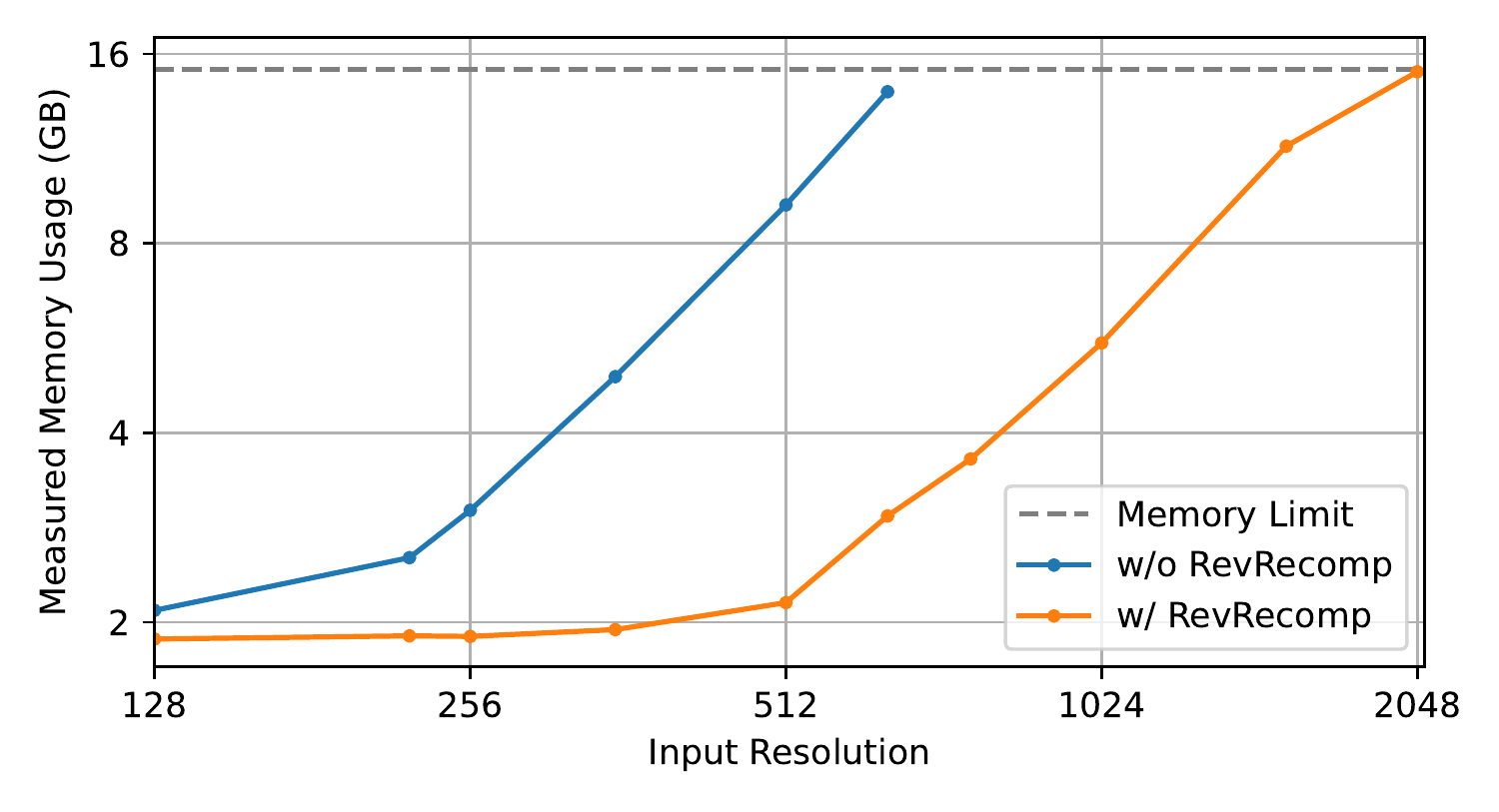}
    \vskip -5pt
    \caption{
        The measured activation memory of training a network using a batch size of 16 on a single GPU with and without reversible recomputation (RevRecomp) as the input resolution is scaled. We observe that with RevRecomp, we can train with higher input resolutions, which is useful in domains such as object detection and segmentation.}
    \label{fig:mvr}
    \end{minipage}
    \vskip -7pt
\end{figure*}

\subsection{Memory Savings}
\label{sec:frbifpn:mem}

The activation memory complexity of training a CV network is $\text{O}(n c h w d)$ where $n$ is the batch size, $c$ is the number of channels representing the network's width, $h$ and $w$ specify the input resolution, and $d$ denotes the depth of the network.
By decoupling depth from activation memory requirements, reversible networks have an activation memory complexity of $\text{O}(n c h w)$.
\cref{fig:mvd} shows the measured memory usage of the RevBiFPN-S0 network as the network depth is scaled with and without reversible recomputation.
This demonstrates that measured memory usage is approximately constant when reversible recomputation is used but increases linearly otherwise.

When scaling width, batch size, or input resolution, networks with and without reversible recomputation have the same complexity, but using reversibility creates a memory offset that enables larger variants to be trained.
As an example, \cref{fig:mvr} shows the measured memory usage of RevBiFPN-S0 as the resolution is varied.
The reversible variant has an advantageous offset and can run resolutions about $4^2$ larger than is possible with a network without reversible recomputation.
On a 16GB system, the largest image a network without reversible recomputation can process is just over 2K$\times$2K.
With reversibility, the same network can process images with resolutions up to 8K$\times$8K.

\subsection{Network Heads}
\label{sec:frbifpn:heads}

While the RevBiFPN backbone is fully reversible, it can be used with non-reversible heads. Before each head is applied, a set of MBConv blocks is used as a neck, with reverse checkpointing, to transform the output channels of RevBiFPN-S0 to $48$, $64$, $128$, and $320$. The dimensionality of the neck and heads is scaled using the width multipliers shown in \cref{tab:scales} (\cref{sec:frbifpn:scaling}).
For the detection and segmentation networks, the input resolution is also modified.

\begin{figure}
    \centering
    \includegraphics[width=\linewidth]{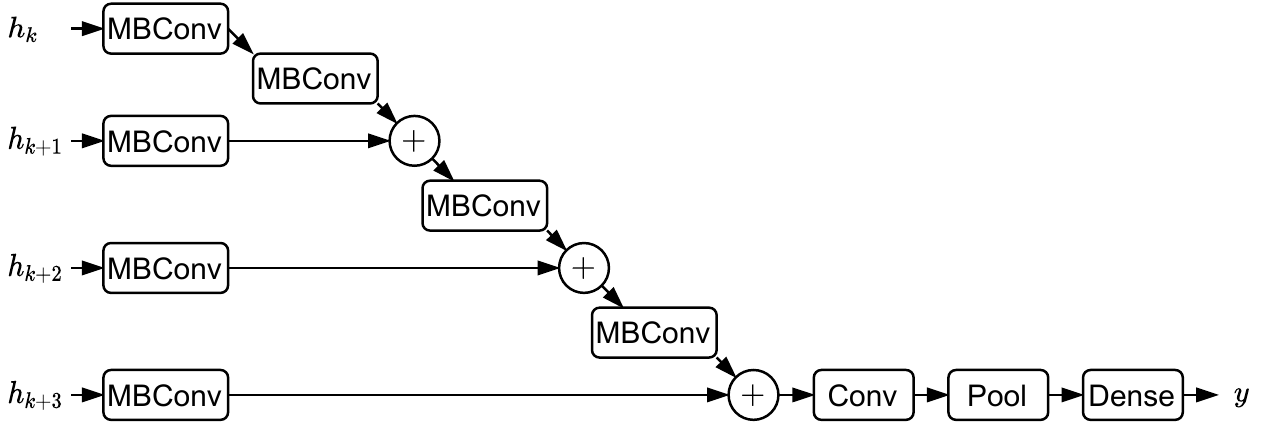}
    \vskip 5pt
    \caption{Neck and classification head with feature pyramid input.}
    \label{fig:c_head}
    \vskip -10pt
\end{figure}

ImageNet classification is used to pretrain the RevBiFPN backbone before it is fine-tuned for object detection and segmentation.
The backbone outputs a feature pyramid transformed by the neck and non-reversible classification head shown in \cref{fig:c_head}.
In the head, the highest resolution feature map is downsampled by a factor of 2 using an MBConv block with stride 2 and is added to the next largest feature map.
This is repeated multiple times until all information is aggregated into the lowest-resolution feature map.
A $1 \times 1$ convolution is applied, followed by global average pooling and a dense layer. This design is inspired by \citet{sun2019hrnet_pose} but uses the MBConv block.

Object detection and instance segmentation are done with the Faster R-CNN and Mask R-CNN heads provided in MMDetection~\cite{chen2019mmdetection}.

\subsection{Network Scaling}
\label{sec:frbifpn:scaling}

\begin{table}
    \caption{
        Network width multiplier ($m_w$), depth ($d$), and input height and width ($h$ and $w$) of RevBiFPN variants trained on ImageNet at different scales.
        Without reversibility, the training setup must be modified to accommodate scales past RevBiFPN-S1.
        Reversibility enables training RevBiFPN-S6 with an activations that are about 24x larger than that of RevBiFPN-S1.\protect\footnotemark
    }
    \vskip 10pt
    \centering
    \small
    \sc
    \begin{tabular}{lrrrrrr}
        \toprule
        Model       & $m_w$ & $d$ & $h$ and $w$ \\ \midrule
        RevBiFPN-S0 & 1     &   2 & 224 \\
        RevBiFPN-S1 & 1.3   &   2 & 256 \\
        RevBiFPN-S2 & 2     &   2 & 256 \\
        RevBiFPN-S3 & 2.7   &   3 & 288 \\
        RevBiFPN-S4 & 4     &   4 & 320 \\
        RevBiFPN-S5 & 5.3   &   4 & 352 \\
        RevBiFPN-S6 & 6.7   &   5 & 352 \\
        \bottomrule
    \end{tabular}
    \label{tab:scales}
    \vskip -10pt
\end{table}
\footnotetext{
Without reversible recomputation, the memory used for activations, $n \times c \times h \times w \times d$, dominates memory usage.
$|\text{RevBiFPN-S6}|\ /\ |\text{RevBiFPN-S1}| = (n \times 6.67c \times 352 \times 352 \times 5) / (n \times 1.33c \times 256 \times 256 \times 2) = 23.7$.
}

Once the baseline network is designed, scaling the input resolution, width, and depth generally results in better performance.
Classically, networks such as VGG~\cite{simonyan2014very} and ResNet~\cite{he2016deep} focus on scaling network depth.
\citet{tan2019efficientnet} shows how compound scaling, i.e., scaling all dimensions, results in efficient networks at all parameters and MAC counts.
\citet{dollar2021fast} shows how to scale such that the network run-time is minimized for large networks.
Equations (4) and (5) of \citet{dollar2021fast} produce a ``family of scaling strategies parameterized by $\alpha$.''

This work uses these scaling strategies but sets $\alpha = 2/3$.
While \citet{dollar2021fast} recommends $\alpha = 4/5$, they also show $\alpha = 2/3$ is nearly as fast but prioritizes depth and resolution scaling.
This gives added memory benefits in the reversible setting (\cref{sec:frbifpn:mem}).
Given the outputs of the scaling strategy, $m_w$ is chosen such that channel counts are multiples of 16, the depth is rounded to the nearest integer, and the resolution is set to a multiple of $2^5$ (\cref{tab:scales}).

\section{Experiments}
\label{sec:exp}


\begin{table}
    \caption{Single-crop, single-model ImageNet accuracy.}
    \vskip 10pt
    \centering
    \small
    \sc
    \begin{tabular}{lrrr}
        \toprule
        Model       &  Params &  MACs &   Top1 \\ \midrule
        RevBiFPN-S0 &   3.42M & 0.31B & 72.8\% \\
        RevBiFPN-S1 &   5.11M & 0.62B & 75.9\% \\
        RevBiFPN-S2 &   10.6M & 1.37B & 79.0\% \\
        RevBiFPN-S3 &   19.6M & 3.33B & 81.1\% \\
        RevBiFPN-S4 &   48.7M & 10.6B & 83.0\% \\
        RevBiFPN-S5 &   82.0M & 21.8B & 83.7\% \\
        RevBiFPN-S6 &  142.3M & 38.1B & 84.2\% \\
        \bottomrule
    \end{tabular}
    \vskip -7pt
    \label{tab:i1kacc}
\end{table}

\subsection{ImageNet Classification}
\label{sec:exp:cls}

\textbf{Setup.} The ImageNet dataset is used to pretrain the network before variants are fine-tuned on downstream tasks.
All RevBiFPN variants are pretrained for $350$ epochs using $8$ GPUs with a per GPU batch size of $64$. This enables higher throughput for training while amortizing the costs of reverisble recomputation.
SGD is used with a learning rate of $0.1$ and momentum of $0.9$, and
an exponential moving average (EMA) of the network parameters is used with a decay of $0.9999$.
A $5$ epoch learning rate warm-up is used with a starting learning rate of $10^{-3}$ followed by cosine decay \cite{loshchilov2017sgdr}.
The last ten epochs use a constant learning rate of $10^{-4}$.
The network uses batch normalization with a momentum of $0.9$ and epsilon of $10^{-3}$.
Training is regularized using label smoothing~\cite{szegedy2016rethinking}, weight decay, dropout~\cite{srivastava2014dropout}, stochastic depth~\cite{huang2016deep}, CutMix~\cite{yun2019cutmix}, mixup~\cite{zhang2018mixup}, and RandAugment~\cite{cubuk2020randaugment}.
Tuning these parameters could result in further improvement (details in \cref{appx:i1k_regularization}).

\textbf{Results.}
Although RevBiFPN-S0 and RevBiFPN-S1 can be trained without reversible recomputation, unless otherwise stated, all of the results are shown for networks trained with reversible recomputation.
\cref{tab:i1kacc} summarizes ImageNet classification results showing top1 accuracy, parameter counts and MACs used at evaluation.
While not designed primarily for classification, RevBiFPN still produces results comparable to classification-specific networks.
RevBiFPN-S6 uses 38.1B MACs and achieves 84.2\% ImageNet Top1 accuracy, making it comparable to EfficientNet-B7, which uses 37B MACs to achieve 84.3\% ImageNet Top1 accuracy. \cref{tab:i1kacc_ext} in the appendix extends \cref{tab:i1kacc} to include comparisons with other networks.

\begin{table}
    \caption{
        Training Memory (GB) used  per sample.
        The training resolution used for RevBiFPN-S6 is $352$ and EfficientNet-B7 is trained with a resolution of $600$.
    }
    \vskip 10pt
    \centering
    \small
    \sc
    \begin{tabular}{lrrr}
        \toprule
        \multirow{2}{*}{MODEL} & \multicolumn{3}{c}{Input Resolution}                                 \\ \cmidrule{2-4}
                              & \multicolumn{1}{r}{Train Res}      & 224           & 384            \\ \midrule
        \textbf{RevBiFPN-S6}  & \multicolumn{1}{r}{\textbf{0.254}} & \textbf{0.086} & \textbf{0.291} \\
        EfficientNet-B7        & \multicolumn{1}{r}{5.047}         & 0.673         & 1.786         \\
        \bottomrule
    \end{tabular}
    \vskip -7pt
    \label{tab:mem}
\end{table}

\begin{table}
    \caption{
        Slowdown of using reversible recomputation. Theoretically reversible recomputation adds a 33\% compute overhead (\cref{tab:mem_comp_complex}). In practice, with more memory, training can be optimized and can use a larger batch size to help efficiency.
    }
    \vskip 10pt
    \centering
    \small
    \sc
    \begin{tabular}{lr}
        \toprule
        Model       & Slowdown \\ \midrule
        RevBiFPN-S0 &  25.02\% \\
        RevBiFPN-S2 &  21.96\% \\
        RevBiFPN-S4 &  15.73\% \\
        RevBiFPN-S6 &  12.73\% \\
        \bottomrule
    \end{tabular}
    \label{tab:revrecomp_slowdowns}
    \vskip -7pt
\end{table}

\cref{tab:mem} shows that the GPU memory usage of RevBiFPN-S6 is a fraction of the memory used by EfficientNet-B7 at their respective training resolutions
and at input resolutions of $224$ and $384$, which are frequently used in CV backbones.

\cref{tab:revrecomp_slowdowns} shows the measured slowdowns from using reversible recomputation for different networks. We see that as the networks get larger, the overheads reduce considerably, making the technique efficient to use at scale.

\subsubsection{Training With Reversibility}
\label{sec:i1k_rev_training}

\begin{figure}
    \centering
    \includegraphics[width=\linewidth]{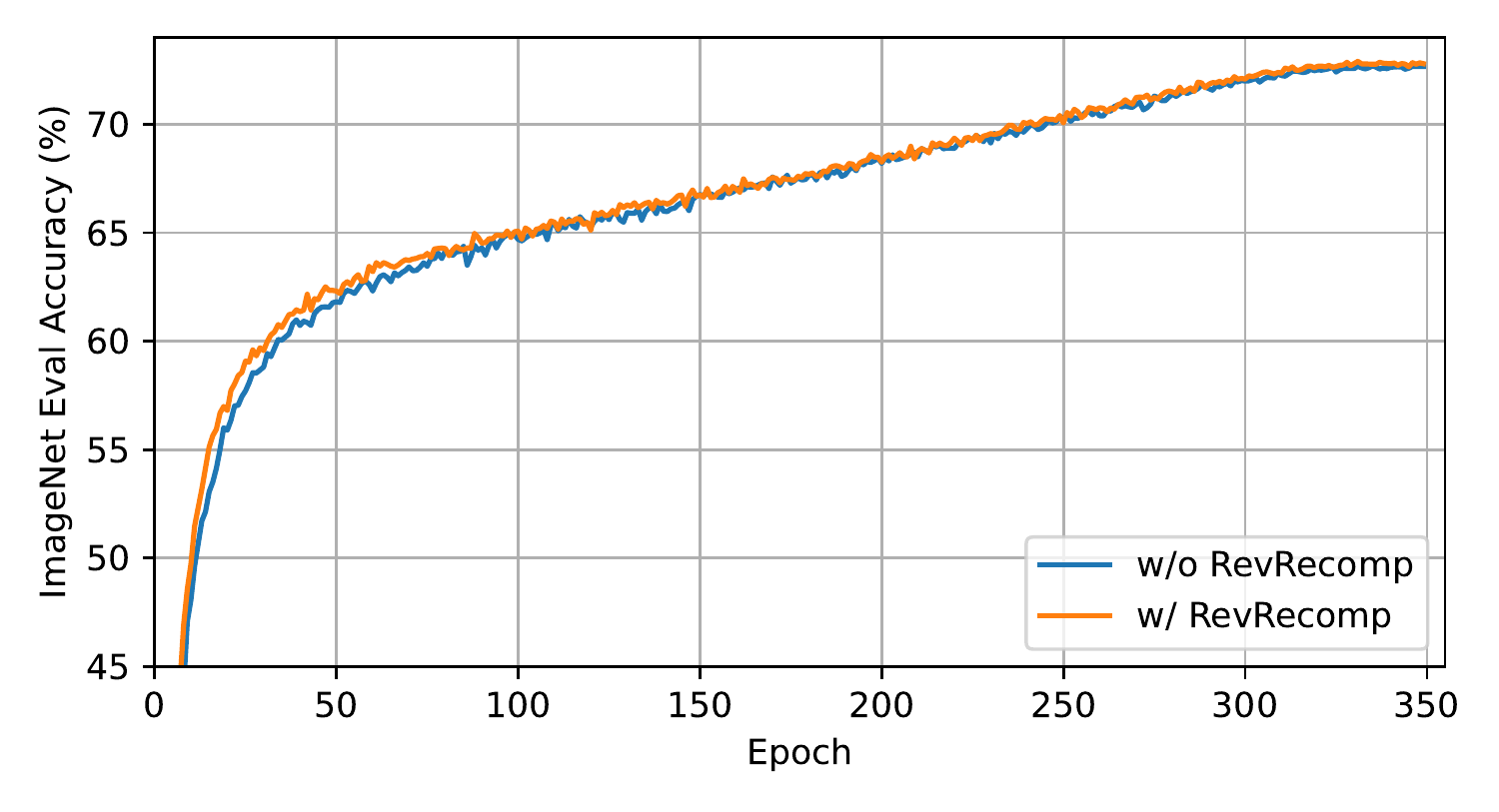}
    \vskip -10pt
    \caption{ImageNet validation accuracy when training RevBiFPN-S0 with and without reversible recomputation (RevRecomp).}
    \label{fig:i1k_rev_training}
    \vskip -10pt
\end{figure}

Under infinite precision, training with and without reversible recomputation would produce identical results.
\cref{fig:i1k_rev_training} shows that while there are differences when using finite precision, these are inconsequential.
Training RevBiFPN-S0 with reversible recomputation requires only 2GB of accelerator memory and produces results nearly indistinguishable from regular training, which consumes 12GB of memory.

\subsubsection{Architectural Ablations}
\label{appx:Ablations}

As noted in \cref{sec:frbifpn}, RevBiFPN is structurally similar to HRNet.
In this section, RevBiFPN is trained on ImageNet for 150 epochs at an input resolution of 96$\times$96 to ablate architectural design decisions.

\begin{table}
    \caption{Down \& Up Sampling Operation's influence on Accuracy.}
        \vskip 10pt
        \centering
        \small
        \sc
        \begin{tabular}{lrrr}
            \toprule
            Down / Up Sampling & Params &  MACs & Top1   \\ \midrule
            LD / SU            & 3.49M  & 75.7M & 61.5\% \\ 
            SD / SU            & 3.28M  & 67.2M & 60.8\% \\ 
            SD / LU            & 3.47M  & 69.5M & 61.5\% \\ 
            \bottomrule
        \end{tabular}
        \vskip -7pt
        \label{tab:i1kacc_dusample}
\end{table}

\textbf{Down and Up Sampling Operation.} HRNet uses $k$ stride 2 convolution blocks to downsample by $2^k$ (LD).
An alternative downsampling schema would use a single block with stride $2^k$ and an increased kernel size such that the entire input is used to produce the output (SD).
HRNet uses a 1$\times$1 convolution paired with an upsample operation in the `nearest' mode (SU) to upsample feature maps.
The 1$\times$1 convolution does not operate in the spatial domain, and the upsample operation is in the `nearest' mode.
This is ablated by changing the upsampling block to use a 3$\times$3 convolution paired with an upsample operation in `bilinear' mode (LU).

While replacing LD with SD curbs accuracy on ImageNet, augmenting this change by replacing SU with LU results in a total MAC decrease of about 8\% while not affecting ImageNet accuracy (\cref{tab:i1kacc_dusample}).

\textbf{Backbone Stem.} Common practice dictates using a convolutional stem for neural network design.
\citet{ridnik2020tresnet} proposes replacing this with the SpaceToDepth stem.
Their work shows this does not affect network accuracy while increasing GPU throughput performance.
\cref{tab:i1kacc_stem} reaffirms their results and highlights the resulting MAC decrease.

\textbf{Squeeze-Excite.} \citet{ridnik2020tresnet} notes that when applied to ``low-resolution maps, Squeeze-Excite does not get a large accuracy benefit from the global average pooling operation that SE provides.''
They advocate for using Squeeze-Excite on large spatial resolutions as opposed to small spatial resolutions, as this provides a good accuracy vs. throughput tradeoff.
\cref{tab:i1kacc_squexcite} affirms their result by showing how Squeeze-Excite, when applied to the low-resolution path, leaves accuracy relatively unaffected, but when applied to the high-resolution path, improves performance.

\begin{table}
    \caption{Stem's influence on Accuracy.}
        \vskip 10pt
        \centering
        \small
        \sc
        \begin{tabular}{lrrr}
            \toprule
            Stem          & Params &  MACs & Top1   \\ \midrule
            Convolutional & 3.49M  & 75.7M & 61.5\% \\ 
            SpaceToDepth  & 3.49M  & 73.7M & 61.5\% \\ 
            \bottomrule
        \end{tabular}
        \vskip -5pt
        \label{tab:i1kacc_stem}
\end{table}

\begin{table}
    \caption{Influence of Squeeze-Excite.}
        \vskip 10pt
        \centering
        \small
        \sc
        \begin{tabular}{lrrr}
            \toprule
            Squeeze-Excite & Params &  MACs & Top1   \\ \midrule
            None           & 3.40M  & 75.5M & 61.3\% \\
            low-res path   & 3.49M  & 75.7M & 61.4\% \\
            high-res path  & 3.46M  & 76.1M & 61.6\% \\
            \bottomrule
        \end{tabular}
        \vskip -5pt
        \label{tab:i1kacc_squexcite}
\end{table}


\begin{figure*}
    \begin{minipage}{0.48\linewidth}
        \centering
        \begin{overpic}[width=\linewidth]
            {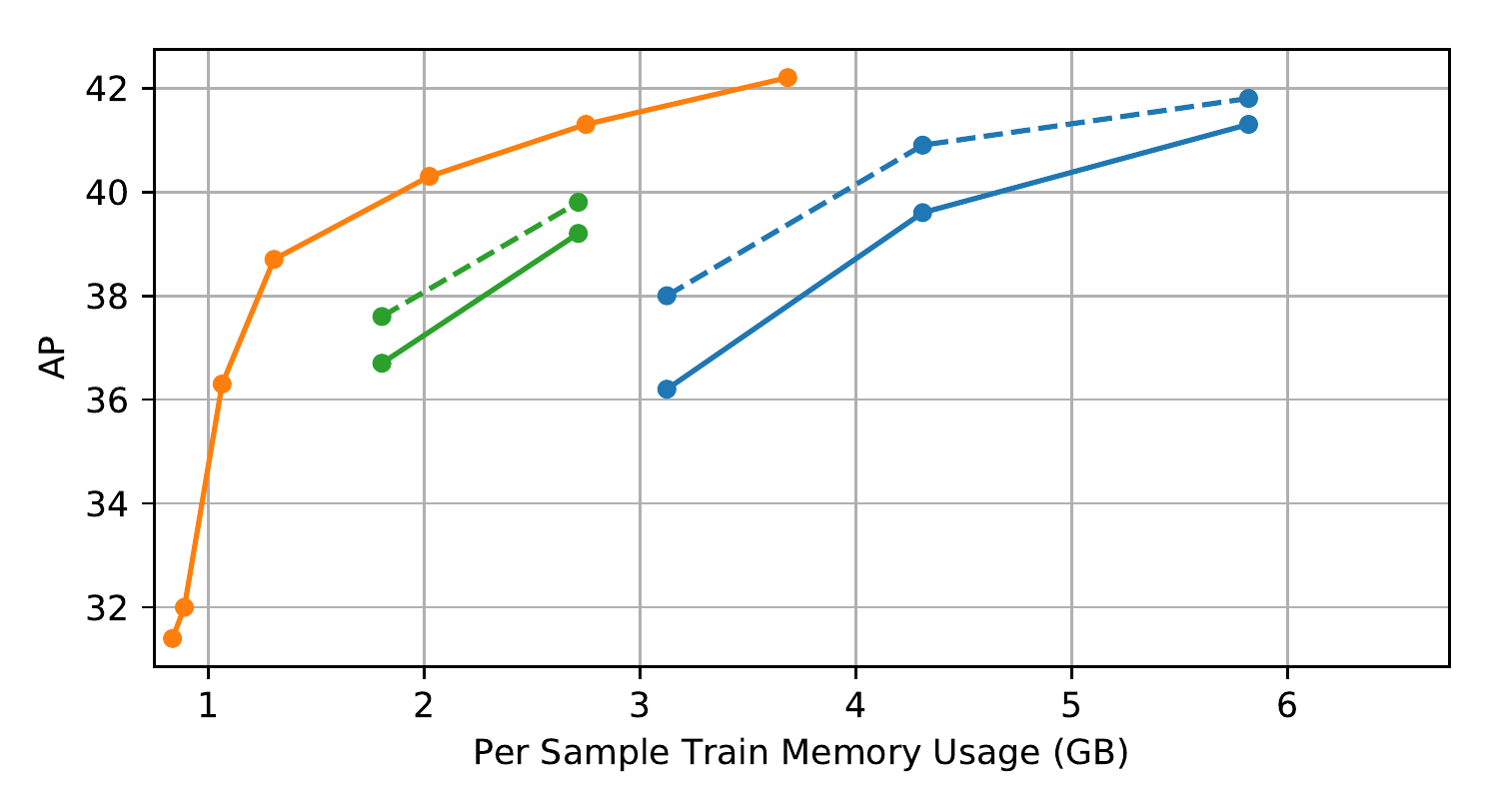}
            \put(50.5,10){ 
                \tcbox[size=minimal,colback=white,colframe=white]{
                \scalebox{0.56}{
                \centering
                \small
                \begin{tabular}{lrrr}
                    \toprule
                    Backbone              &   MACs &     Mem & AP   \\ \midrule
                    RevBiFPN-S3           & 181.0B & 1.31 GB & 38.7 \\
                    RevBiFPN-S5           & 328.9B & 2.75 GB & 41.3 \\ 
                    RevBiFPN-S6           & 465.4B & 3.69 GB & 42.2 \\ \midrule
                    HRNetV2p-W18          & 196.2B & 3.13 GB & 36.2 \\
                    HRNetV2p-W32          & 299.0B & 4.31 GB & 39.6 \\ 
                    HRNetV2p-W48          & 481.9B & 5.82 GB & 41.3 \\ 
                    ResNet-101-FPN        & 296.6B & 2.72 GB & 39.2 \\
                    \bottomrule
                \end{tabular}
            }}}
            \put(43,25){ 
                \tcbox[size=minimal,colback=white,colframe=white]{
                \scalebox{0.45}{\textsf{W18}}
            }}
            \put(60,37){ 
                \tcbox[size=minimal,colback=white,colframe=white]{
                \scalebox{0.45}{\textsf{W32}}
            }}
            \put(76,41){ 
                \tcbox[size=minimal,colback=white,colframe=white]{
                \scalebox{0.45}{\textsf{HRNetV2p-W48}}
            }}
            \put(22,27){ 
                \tcbox[size=minimal,colback=white,colframe=white]{
                \scalebox{0.45}{\textsf{ResNet-50-FPN}}
            }}
            \put(32,35.5){ 
                \tcbox[size=minimal,colback=white,colframe=white]{
                \scalebox{0.45}{\textsf{101-FPN}}
            }}
            \put(11.25,10){ 
                \tcbox[size=minimal,colback=white,colframe=white]{
                \scalebox{0.45}{\textsf{S0}}
            }}
            \put(12,13){ 
                \tcbox[size=minimal,colback=white,colframe=white]{
                \scalebox{0.45}{\textsf{S1}}
            }}
            \put(15,27){ 
                \tcbox[size=minimal,colback=white,colframe=white]{
                \scalebox{0.45}{\textsf{S2}}
            }}
            \put(18,35){ 
                \tcbox[size=minimal,colback=white,colframe=white]{
                \scalebox{0.45}{\textsf{S3}}
            }}
            \put(28,40){ 
                \tcbox[size=minimal,colback=white,colframe=white]{
                \scalebox{0.45}{\textsf{S4}}
            }}
            \put(38,43){ 
                \tcbox[size=minimal,colback=white,colframe=white]{
                \scalebox{0.45}{\textsf{S5}}
            }}
            \put(50,46.3){ 
                \tcbox[size=minimal,colback=white,colframe=white]{
                \scalebox{0.45}{\textsf{RevBiFPN-S6}}
            }}
        \end{overpic}
        \vskip -7pt
        \caption{
            Object detection results on COCO \texttt{minival} in the Faster R-CNN framework as a function of memory used for training.
            Tables show results for 1x schedule.
        }
        \label{fig:memvsap_faster}
    \end{minipage}
    \hfill
    \begin{minipage}{0.48\linewidth}
        \centering
        \begin{overpic}[width=\linewidth]
            {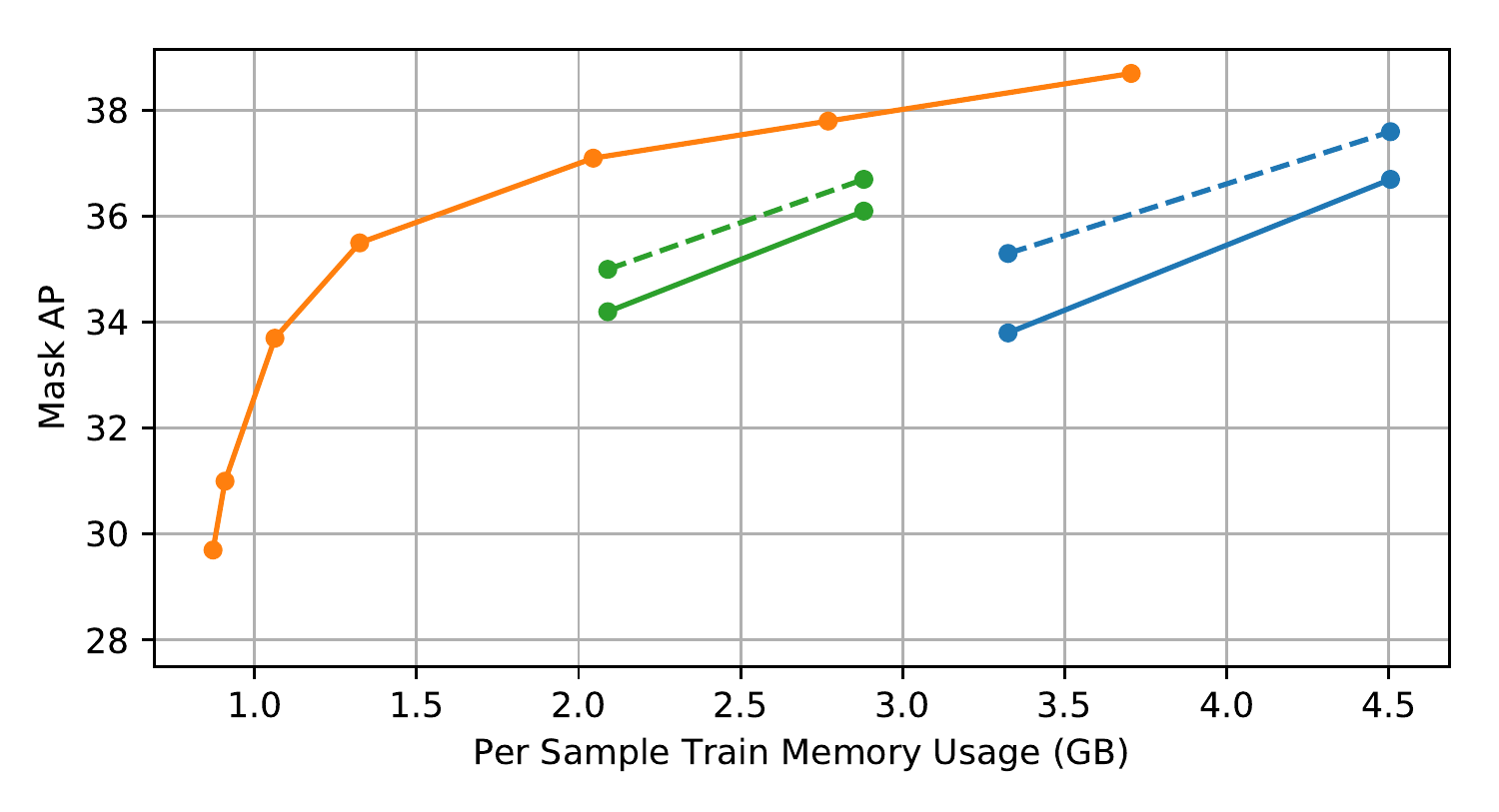}
            \put(49.5,10){ 
                \tcbox[size=minimal,colback=white,colframe=white]{
                \scalebox{0.56}{
                \centering
                \small
                \begin{tabular}{lrrr}
                    \toprule
                    Backbone       &    MACs &     Mem &   AP \\ \midrule
                    RevBiFPN-S2   & 210.49B & 1.06 GB & 33.7 \\
                    RevBiFPN-S4   & 304.09B & 2.05 GB & 37.1 \\ 
                    RevBiFPN-S6   & 518.50B & 3.71 GB & 38.7 \\ \midrule
                    HRNetV2p-W18   & 249.25B & 3.33 GB & 33.8 \\
                    HRNetV2p-W32   & 352.03B & 4.51 GB & 36.7 \\ 
                    ResNet-101-FPN & 349.65B & 2.88 GB & 36.1 \\
                    \bottomrule
                \end{tabular}
            }}}
            \put(69.25,31){ 
                \tcbox[size=minimal,colback=white,colframe=white]{
                \scalebox{0.45}{\textsf{HRNetV2p-W18}}
            }}
            \put(90,39.4){ 
                \tcbox[size=minimal,colback=white,colframe=white]{
                \scalebox{0.45}{\textsf{W32}}
            }}
            \put(31,30){ 
                \tcbox[size=minimal,colback=white,colframe=white]{
                \scalebox{0.45}{\textsf{ResNet-50-FPN}}
            }}
            \put(53,37){ 
                \tcbox[size=minimal,colback=white,colframe=white]{
                \scalebox{0.45}{\textsf{101-FPN}}
            }}
            \put(14,16){ 
                \tcbox[size=minimal,colback=white,colframe=white]{
                \scalebox{0.45}{\textsf{S0}}
            }}
            \put(14.5,20){ 
                \tcbox[size=minimal,colback=white,colframe=white]{
                \scalebox{0.45}{\textsf{S1}}
            }}
            \put(18,30){ 
                \tcbox[size=minimal,colback=white,colframe=white]{
                \scalebox{0.45}{\textsf{S2}}
            }}
            \put(23,35.75){ 
                \tcbox[size=minimal,colback=white,colframe=white]{
                \scalebox{0.45}{\textsf{S3}}
            }}
            \put(39,41){ 
                \tcbox[size=minimal,colback=white,colframe=white]{
                \scalebox{0.45}{\textsf{S4}}
            }}
            \put(54.5,43.5){ 
                \tcbox[size=minimal,colback=white,colframe=white]{
                \scalebox{0.45}{\textsf{S5}}
            }}
            \put(72,46.5){ 
                \tcbox[size=minimal,colback=white,colframe=white]{
                \scalebox{0.45}{\textsf{RevBiFPN-S6}}
            }}
        \end{overpic}
        \vskip -7pt
        \caption{
            Instance segmentation results on COCO \texttt{minival} in the Mask R-CNN framework as a function of memory used for training.
            Tables show results for 1x schedule.
        }
        \label{fig:memvsap_mask}
    \end{minipage}
    \vskip 10pt

    \begin{minipage}{0.48\linewidth}
        \centering
        \includegraphics[width=\linewidth]{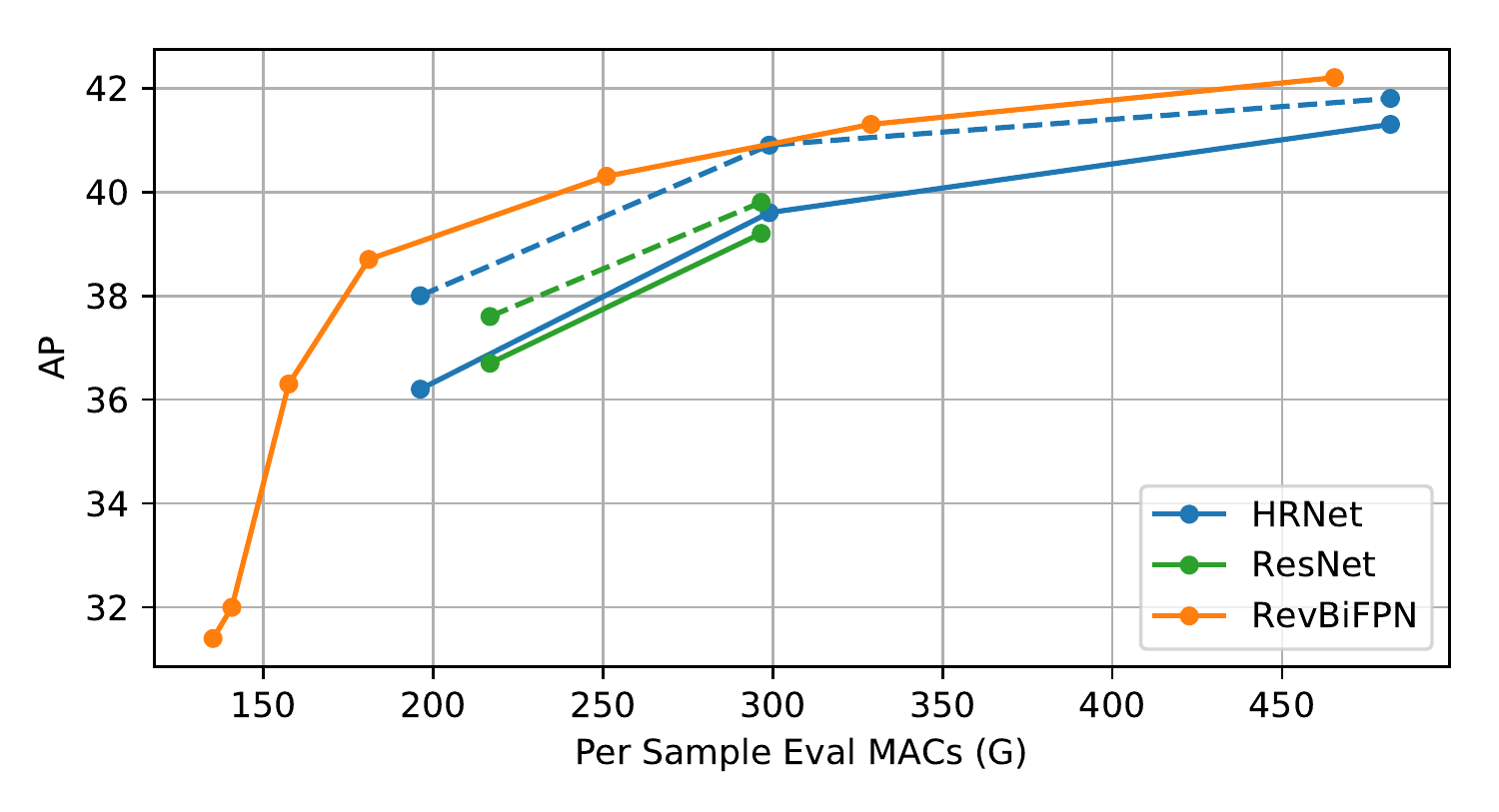}
        \vskip -10pt
        \caption{
            Object detection results on COCO \texttt{minival} in the Faster R-CNN framework as a function of evaluation MACs.
        }
        \label{fig:macvsap_faster}
    \end{minipage}
    \hfill
    \begin{minipage}{0.48\linewidth}
        \centering
        \includegraphics[width=\linewidth]{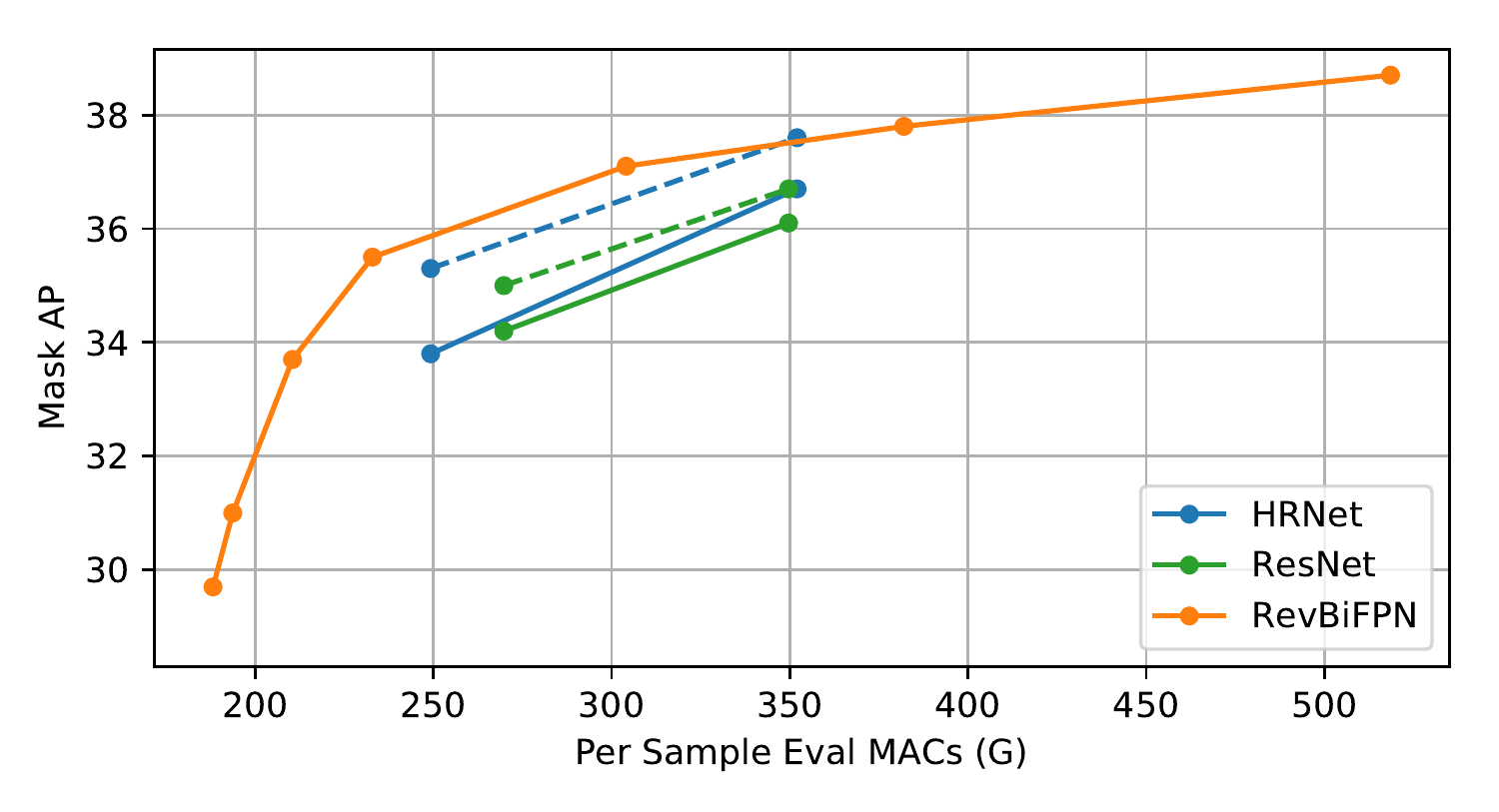}
        \vskip -10pt
        \caption{
            Instance segmentation results on COCO \texttt{minival} in the Mask R-CNN framework as a function of evaluation MACs.
        }
        \label{fig:macvsap_mask}
    \end{minipage}
\end{figure*}

\subsection{MSCOCO Object Detection and Instance Segmentation}
\label{sec:exp:coco}

\textbf{Setup.} Experimental results are presented on the MS COCO 2017 detection dataset, which contains about 118k images for training and 5k for validation ($\texttt{minival}$).
The average precision (AP) metric is adopted, which is the standard COCO evaluation procedure.
The multi-level feature representations from RevBiFPN, as shown in \cref{fig:frbifpn}, are applied for object detection.
There is no additional data augmentation besides standard horizontal flipping.
For training and testing, the input images are resized so that the shorter edge is 800 pixels \cite{lin2017feature}.
Evaluation is performed using a single image scale.

RevBiFPN is compared with HRNet and ResNet.
The object detection performance is evaluated on COCO \texttt{minival} under the two-stage anchor-based framework, Faster-RCNN~\cite{ren2015faster}.
Faster R-CNN models are trained using RevBiFPN, HRNet, and ResNet-FPN as pretrained backbones on the MMDetection open-source object detection toolbox \cite{chen2019mmdetection} using the provided training configurations. 
\cref{fig:memvsap_faster,fig:memvsap_mask,fig:macvsap_faster,fig:macvsap_mask} summarizes the results.
\cref{tab:cocofrcnn,tab:cocomrcnn} in \cref{sec:exp:mscoco_ext_results} extend these results to show parameters, the evaluation MACs per sample,
the GPU memory usage during training, and scores.
Similar to \citet{sun2019hrnet_pose}, the GPU memory usage is measured during training on a 4 GPU system, with an input size of 800$\times$1333 and batch size of 8.

Baseline results are from \citet{wang2020hrnet}.
RevBiFPN is pretrained for 350 epochs whereas \citet{wang2020hrnet} pretrains for 100 epochs.
\citet{he2019rethinking} shows how longer fine-tuning schedules can eliminate the benefits of pretraining.
Although not ideal, this provides a way to compare networks trained for 100 epochs and fine-tuned with a 2x schedule to networks trained for 350 epochs and fine-tuned with a 1x schedule.
Even though most works would only compare networks fine-tuned with the same schedules, 1x and 2x schedules are included to enable such comparisons. 
\citet{tan2020efficientdet} fine-tunes networks for up to 600 epochs.
As a result, EfficientDet~\cite{tan2020efficientdet} serves as a strong baseline for work pursuing SOTA results.
Being subject to resource constraints, we do not make such comparisons and instead focus on memory saving. However, note that \citet{tan2020efficientdet} shows how longer training schedules further differentiate networks using bidirectional multi-scale feature fusion.

\begin{figure*}
    \begin{minipage}{0.48\linewidth}
        \centering
        \includegraphics[width=\linewidth]{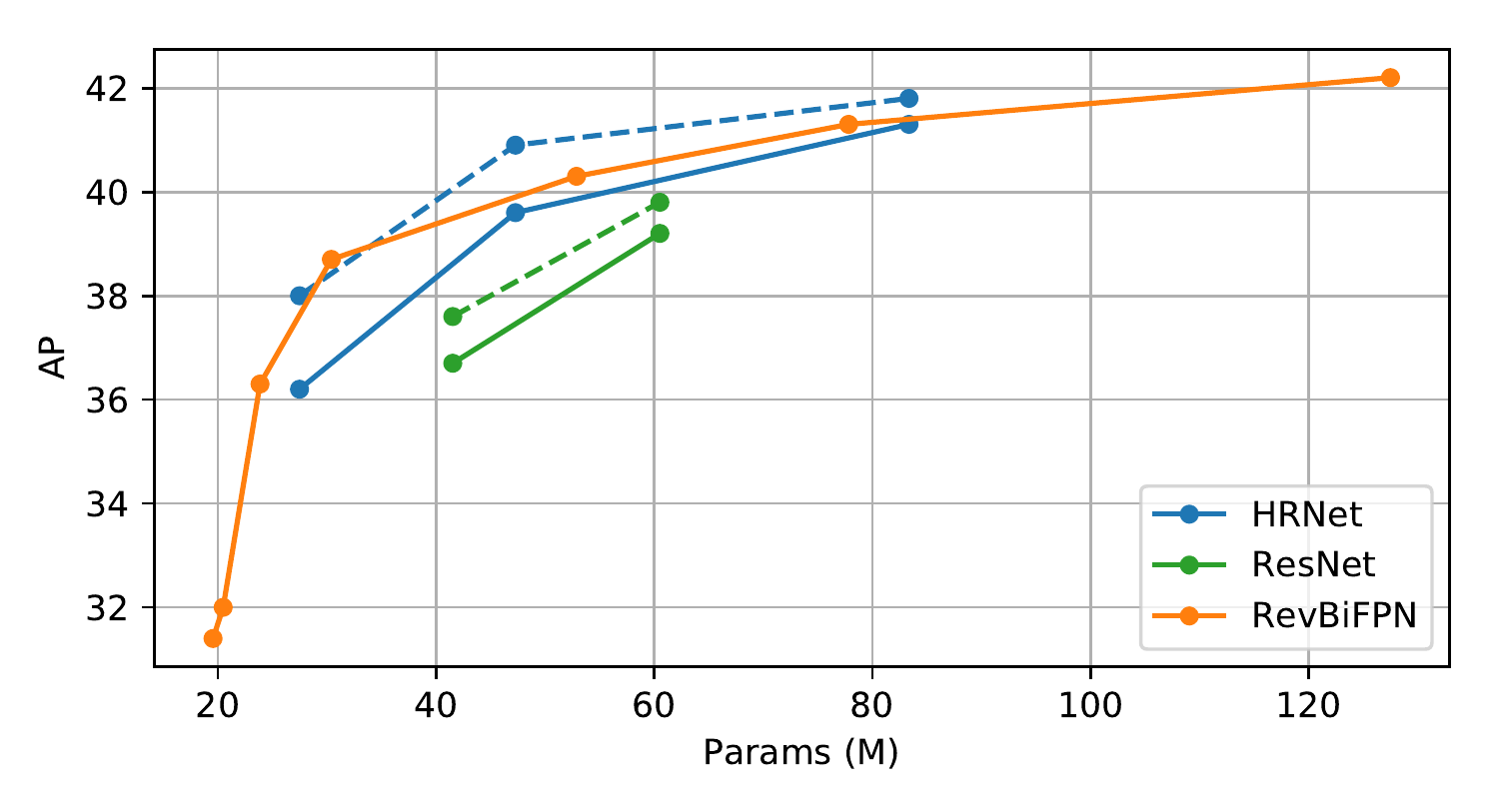}
        \vskip -10pt
        \caption{
            Object detection results on COCO \texttt{minival} in the Faster R-CNN framework as a function of network parameters.
        }
        \label{fig:paramvsap_faster}
    \end{minipage}
    \hfill
    \begin{minipage}{0.48\linewidth}
        \centering
        \includegraphics[width=\linewidth]{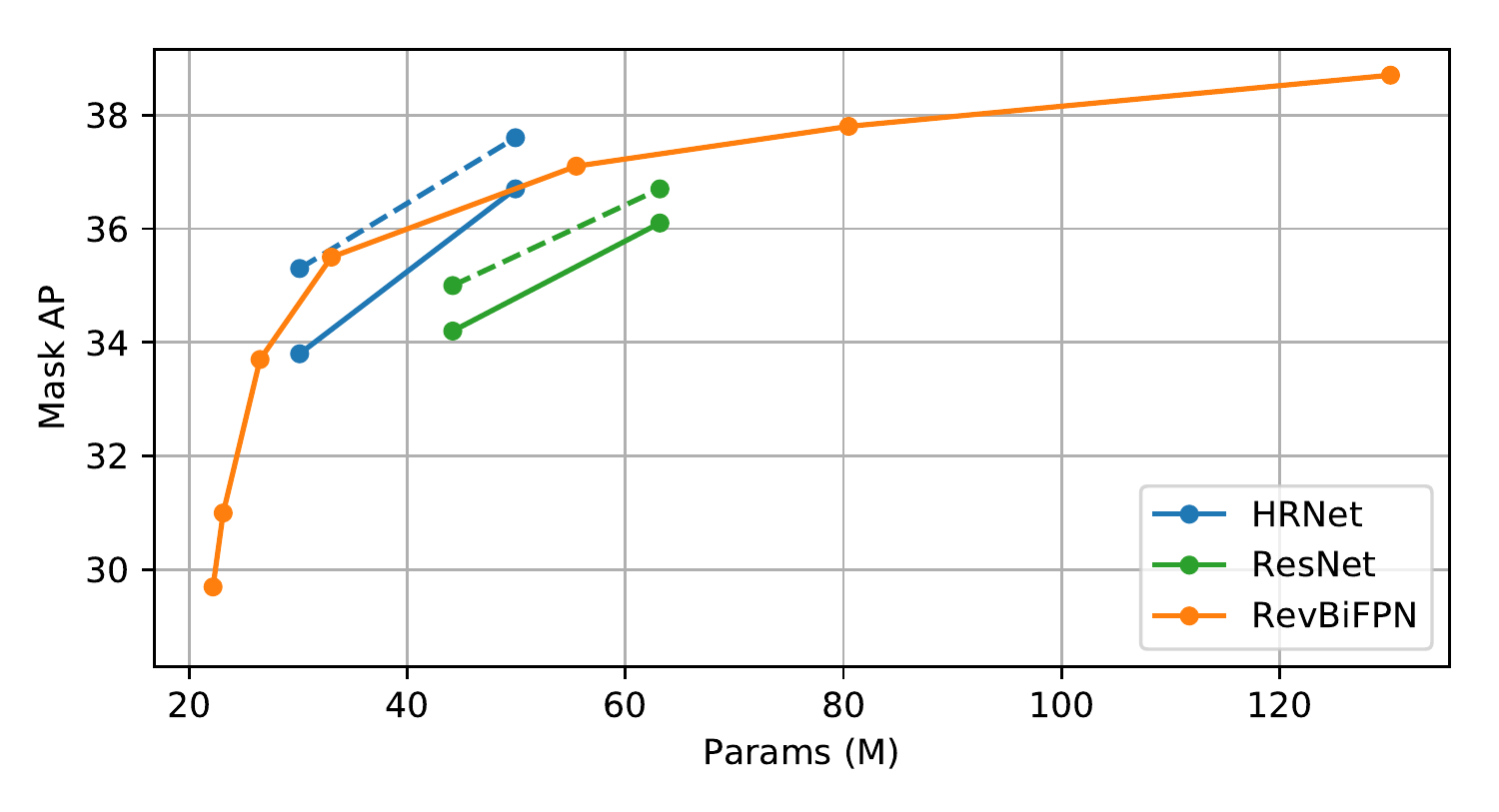}
        \vskip -10pt
        \caption{
            Instance segmentation results on COCO \texttt{minival} in the Mask R-CNN framework as a function of network parameters.
        }
        \label{fig:paramvsap_mask}
    \end{minipage}
\end{figure*}

\textbf{Results.}
\cref{fig:memvsap_faster,fig:memvsap_mask,fig:macvsap_faster,fig:macvsap_mask} show how RevBiFPN, with a 1x fine-tuning schedule, uses less memory and compute\footnote{The tool used to analyze the evaluation MACs per sample for the various models can be found here: \url{https://github.com/open-mmlab/mmcv/blob/master/mmcv/cnn/utils/flops_counter.py}} than the baseline networks at different network performance levels even when those networks are fine-tuned using a 2x schedule.
In \cref{fig:memvsap_faster,fig:memvsap_mask,fig:macvsap_faster,fig:macvsap_mask,fig:paramvsap_faster,fig:paramvsap_mask}, networks fine-tuning using a 1x are shown using solid lines, networks fine-tuning using a 2x are shown using dashed lines.
RevBiFPN is fine-tuned using the HRNet training configurations.
Tuning the hyperparameters could further improve these results.

The Faster R-CNN~\cite{he2017mask} framework is used to evaluate RevBiFPN for object detection on MS COCO.
The results are obtained on the MMDetection toolbox and are summarized in \cref{fig:memvsap_faster,fig:macvsap_faster} and are extended in \cref{tab:cocofrcnn}. 
\cref{fig:memvsap_faster} shows that RevBiFPN-S5 achieves an absolute gain of 3.3\% in AP over HRNetV2p-W18 fine-tuned using the 2x schedule while uses 0.75GB less memory.
RevBiFPN-S3 achieves an absolute gain of 2.5\% in AP over HRNetV2p-W18 using fewer MACs and a \textasciitilde2.4x reduction in training-time memory usage and still outperforms HRNetV2p-W18 by 0.7\% AP even if it is tuned using a 2x schedule.
When the scaled HRNetV2p-W48 is tuned using a 2x schedule, it uses \textasciitilde1.6x the memory and still does not outperform the RevBiFPN-S6 variant tuned using the 1x schedule.

The Mask R-CNN~\cite{he2017mask} framework is used to evaluate RevBiFPN for object detection and instance segmentation on MS COCO.
Results are obtained using the MMDetection toolbox and are summarized in \cref{fig:memvsap_mask,fig:macvsap_mask} and are extended in \cref{tab:cocomrcnn}. 
The overall performance of RevBiFPN-S2 is comparable to HRNetV2p-W18 but uses \textasciitilde1.2x fewer MACs and \textasciitilde2.5x less GPU memory during training. 
RevBiFPN-S6 outperforms HRNetV2p-W32 by 2\% Mask AP and 2.4\% Bbox AP while using 1.6GB less memory and still outperforms HRNetV2p-W32 when it is fine-tuned using the 2x schedule.
Generally, RevBiFPN enables larger batch sizes and image resolutions for detection and segmentation.

\cref{fig:paramvsap_faster,fig:paramvsap_mask} plot the performance of RevBiFPN as a function of network parameters, comparing it to ResNet-FPN and HRNet baselines.
When fine-tuned with a 1x schedule, RevBiFPN outperforms the baseline networks.
HRNet is only able to, per parameter, outperform RevBiFPN when fine-tuned with a 2x schedule (compared to RevBiFPN fine-tuned with a 1x schedule).
While HRNet can be competitive when using parameter count as a metric, \citet{dehghani2021efficiency} and \citet{mehta2021mobilevit} show that network parameter counts produce misleading notions of efficiency and should generally not be used for comparing networks.
The efficiency of RevBiFPN is better communicated by \cref{fig:macvsap_faster,fig:macvsap_mask} where \cref{fig:memvsap_faster,fig:memvsap_mask} shows the memory saving provided by RevBiFPN.

\section{Conclusion}
\label{sec:conclusion}
Bidirectional multi-scale feature fusion has driven progress in computer vision, but accelerator memory often limits network scale.
Reversible methods decrease the activation memory complexity with respect to the depth from linear to constant but were previously not applicable to bidirectional multi-scale feature fusion.
This work introduces RevSilo, a reversible bidirectional multi-scale feature fusion module.
This enables the training of BiFPN-style networks without storing activations.
The RevSilo is used to design RevBiFPN, which is competitive on classification, segmentation, and detection tasks, all while using a fraction of the memory for training.
RevBiFPN applies to memory-constrained settings such as high-resolution detection and segmentation and enables SOTA research without needing hardware with the latest memory capacity.

\subsection{Future Work}
\label{sec:conclusion:futurework}

RevBiFPN is developed to efficiently train networks that drive semantic coherence across feature scales for tasks such as object detection and semantic segmentation.
Tasks such as human pose estimation~\cite{newell2016stacked} and large-scale medical segmentation~\cite{zhou2018unet} can also benefit from such efficiency gains.
\citet{awiszus2020toad} argue that multi-scale processing is needed to generate images using Generative Adversarial Networks (GANs). 
Prior to this work, bidirectional multi-scale feature fusion wasn't possible in flow models, but now the RevSilo and RevBiFPN enable multi-scale fusions in flow-based generation.

While we focus on developing methods and networks for efficient training, tuning models for inference efficiency is also an active direction of research~\cite{mehta2021mobilevit,sandler2018mobilenetv2,ding2021repvgg}. 
To motivate using alternate hardware accelerators, such as the Cerebras Wafer Scale Engine~\cite{liehotchips,cerebrasHarnessingPower}, we use MBConv as the primary building block of the network.
However, based on the target inference device, one can also swap out the basic building block to the ResNet~\cite{he2016deep} or Transformer block~\cite{vaswani2017attention}.
Also, methods such as pruning~\cite{han2015learning,yu2017scalpel} and quantization~\cite{han2015deep,krishnamoorthi2018quantizing} are standard for supporting real-world inference scenarios. 
The impact of reversible recomputation on such methods has yet to be explored in the context of the RevSilo.
Finally, while we rely on Fast Scaling~\cite{dollar2021fast} for our models, researchers can use alternative strategies to scaling models such as Neural Architecture Search~\cite{tan2019mnasnet, ghiasi2019nasfpn, tan2019efficientnet}, while adding memory constraints~\cite{zhao2021memory} to derive model configurations that satisfy inference requirements.

\section*{Acknowledgements}

We thank Shreyas Saxena, Nolan Dey, and Valentina Popescu for their help and comments that improved the manuscript. We also thank Ben Wang, Ross Wightman, Lucas Nestler, Jan XMaster, Alexander Mattick, Atli Kosson, Abhinav Venigalla, Xin Wang, Vinay Rao, and Kenyon (Chuan-Yung) Tsai for insightful discussions.

\clearpage

\bibliography{revbifpn}
\bibliographystyle{mlsys2023}

{
\transparent{0.0} \lipsum[1-3]}

\appendix


\begin{figure*}[!ht]
    \centering
    \includegraphics[width=\textwidth]{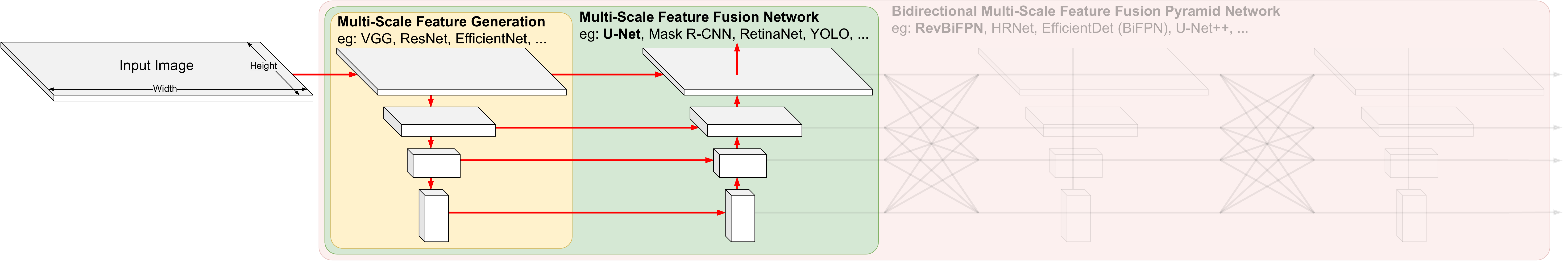}
    \vskip 10pt
    
    \includegraphics[width=\textwidth]{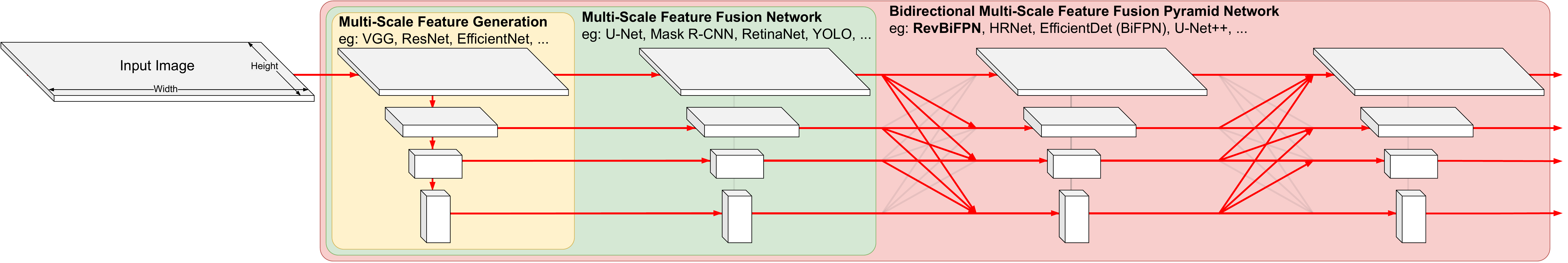}
    \vskip -5pt
    \caption{Effective multi-scale connectivity of U-Net (Top) and RevBiFPN (Bottom).}
    \label{fig:bifpn:effective}
\end{figure*}

\newpage
\section{Examples Multi-Scale Connectivity}
\label{sec:cnnectivity_ex}

\cref{fig:bifpn} encapsulates the connectivity of many network types.
Neural Networks have directed connectivity where the network instantiation dictates the direction of the flow of information and is therefore ambiguous unless pertaining to a particular network instance.
With an instantiation, the connectivity can be directed.
\cref{fig:bifpn:effective} show two examples of instantiated connectivity highlighted in red. The connectivity in the red box can be iteratively applied to further bridge the semantic gap between consecutive feature blocks. For example, EfficientDet~\cite{tan2020efficientdet} can be formed by adding all-to-all connectivity across each feature scale and block.

\section{Reversible Stacked Hourglass Networks}

The reversible residual block \cite{gomez2017reversible} has only been applied to networks with constant hidden dimensionality.
Stacked Hourglass~\cite{newell2016stacked} networks are built using a stack of hourglass structures that maintain constant dimensionality.
Placing each hourglass structure inside a reversible residual block allows the network to produce high-resolution feature maps without storing hidden activations.
To enable comparisons with RevBiFPN variants, we implement a fully reversible Stacked Hourglass Network, RevSHNet.
RevSHNet uses the MBConv block, a SpaceToDepth stem, channel counts similar to RevBiFPN-S0 channel counts, and a comparable classification head.

\subsection{Memory}
\label{appx:rev_sh:mem}

\begin{figure}
    \centering
    \vskip -10pt
    \includegraphics[width=\linewidth]{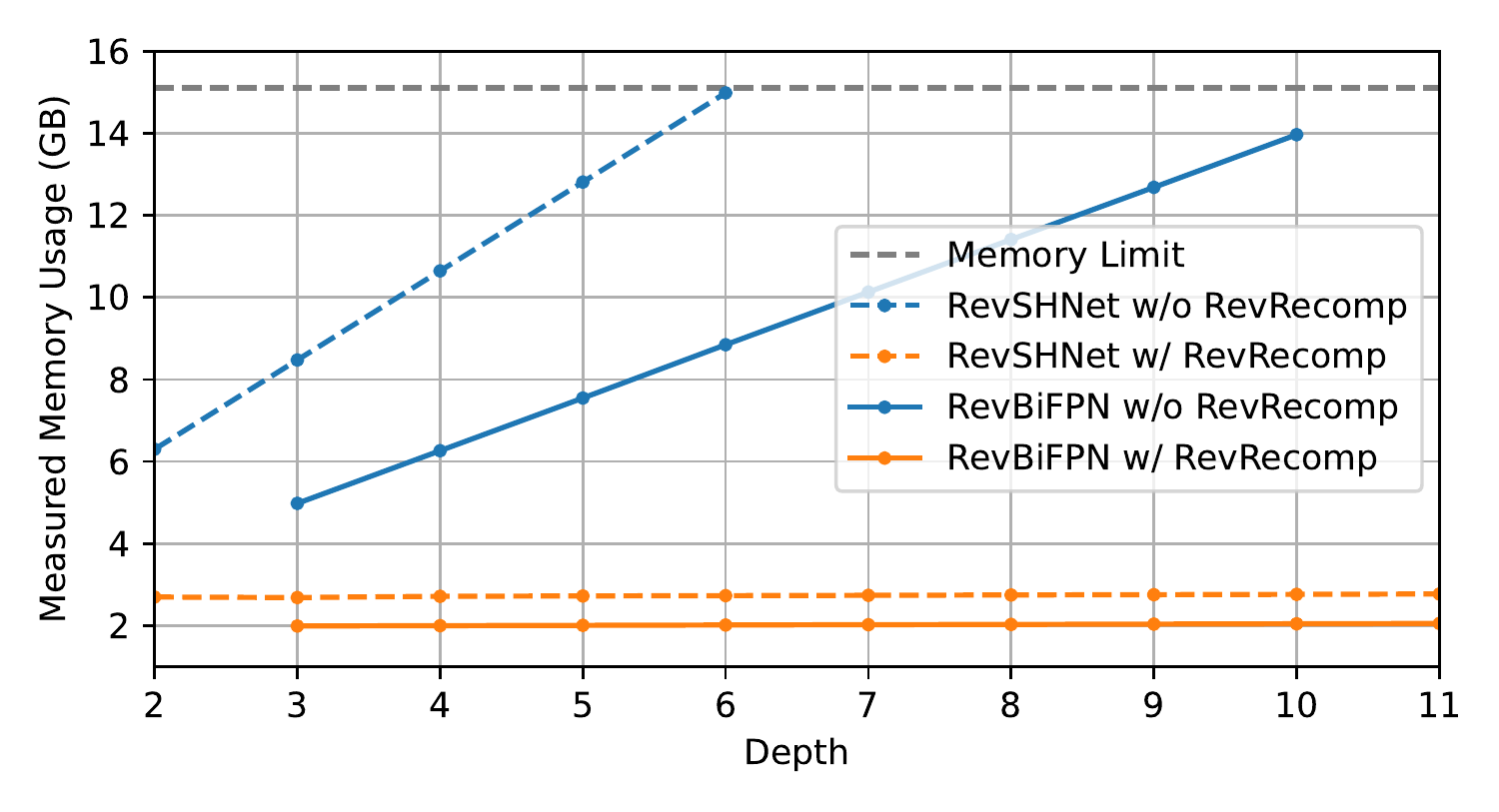}
    \vskip -10pt
    \caption{
        Memory used by RevBiFPN and RevSHNet as depth is scaled with and without reversible recomputation (RevRecomp).}
    \label{fig:rev_sh:mem}
    \vskip -10pt
\end{figure}

Even with reversible recomputation enabled, RevSHNet needs to store an entire hourglass of activations. An input size of 224 results in a memory usage increase of about 40\% when compared to RevBiFPN (\cref{fig:rev_sh:mem}).

\begin{figure}
    \centering
    \includegraphics[width=\linewidth]{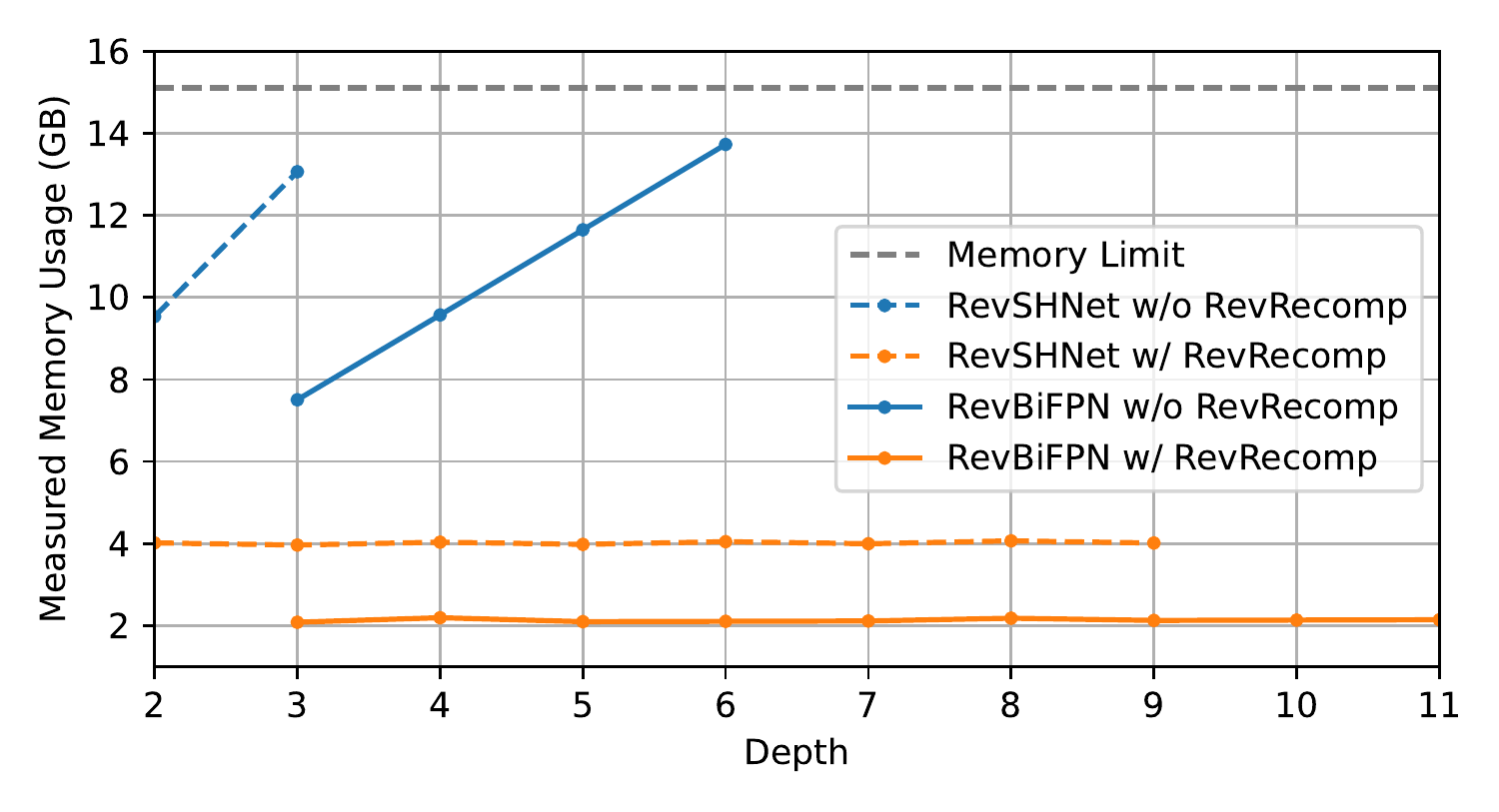}
    \vskip -10pt
    \caption{
        Recreates \cref{fig:rev_sh:mem} with an input resolution of 288.
        RevBiFPN becomes more favorable as resolution is scaled.}
    \label{fig:rev_sh:mem288}
    \vskip -10pt
\end{figure}

\subsection{Compute Complexity}
\label{appx:rev_sh:FLOP}

When the input size is increased to 288, RevSHNet uses almost twice the memory used by RevBiFPN (\cref{fig:rev_sh:mem288}). The increased memory usage limits memory savings and how much the network can be scaled.

\begin{figure}
    \centering
    \includegraphics[width=\linewidth]{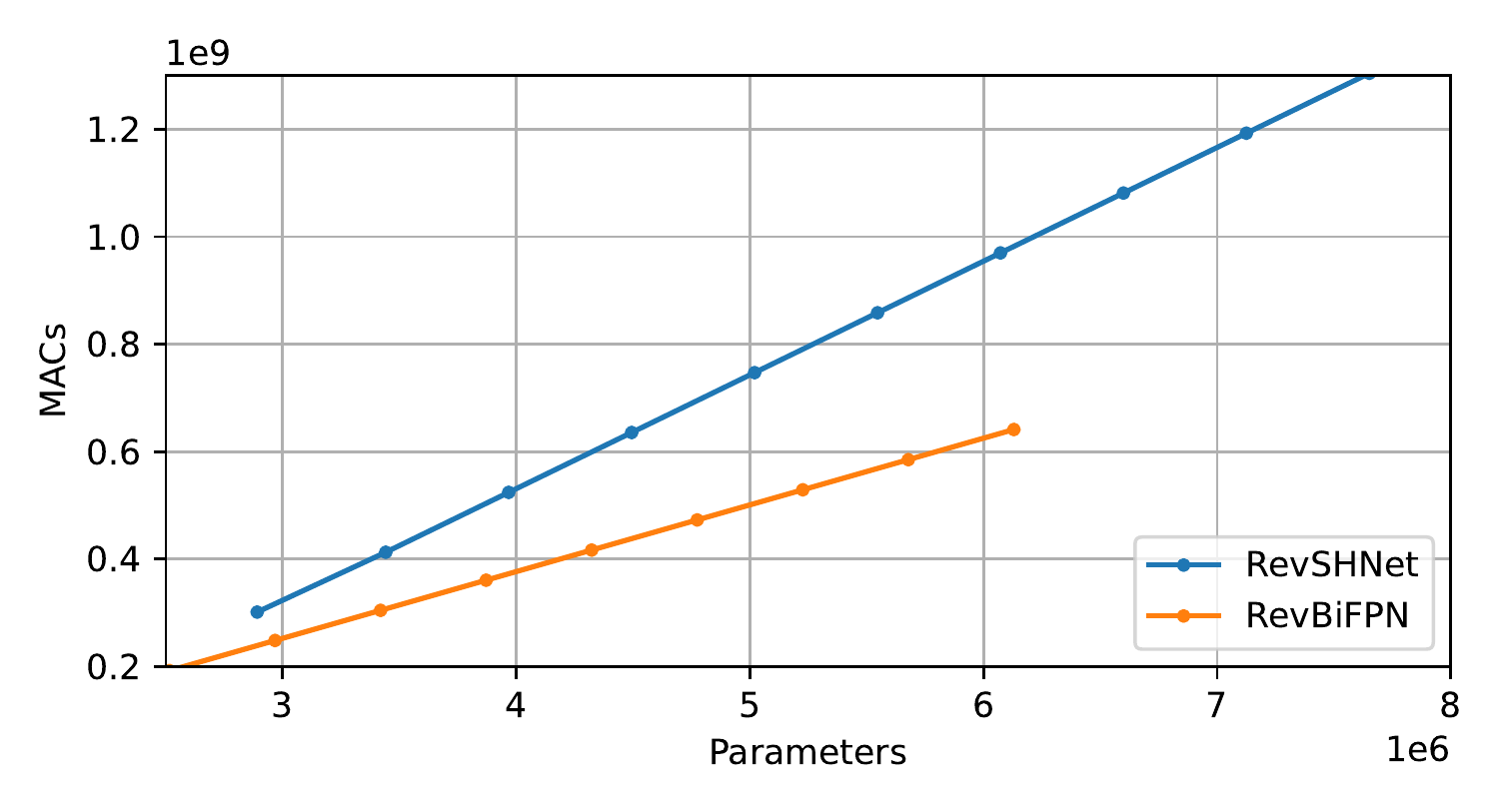}
    \vskip -10pt
    \caption{MACs vs Parameter count of RevBiFPN and RevSHNet as depth is scaled.}
    \label{fig:rev_sh_mac_v_p}
\end{figure}

When RevSHNet is scaled, the produced network has a high compute complexity (\cref{fig:rev_sh_mac_v_p}). This is potentially not optimal and wasteful when scaling networks to larger sizes. It should be noted that the above analysis does not consider network performance. Given comparable networks, we expect RevBiFPN to outperform RevSHNet since RevBiFPN has full bidirectional multi-scale feature fusion with a feature pyramid output, whereas RevSHNet does not.

\section{RevSilo with Additive Coupling}
\label{appx:a_revsilo}

\begin{figure}[h]
    \centering
    \includegraphics[width=0.9\linewidth]{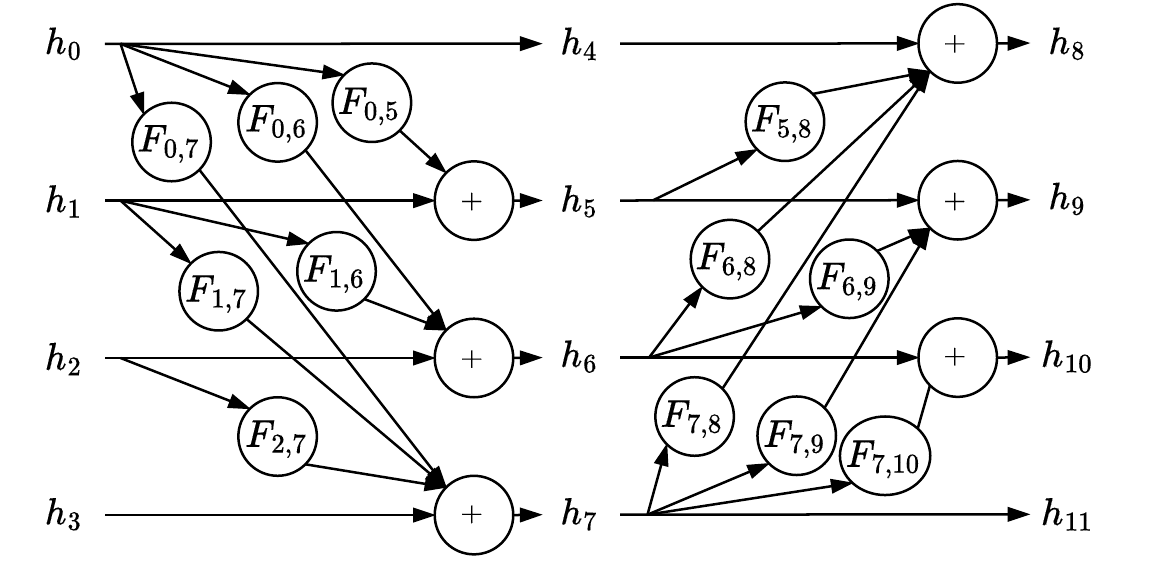}
    \caption{A RevSilo using additive coupling.}
    \label{fig:a_silo}
\end{figure}

While $g_j$ can be any invertible coupling function,
as shown in \cref{fig:a_silo}
this work uses additive coupling and $F_j=\sum_i F_{i,j}\left(x_i\right)$.

For instance \cref{eqn:h7} is computed as:
\begin{equation}
    h_{7} = h_{3} + \left(F_{2,7}\left(h_2\right) + F_{1,7}\left(h_1\right) + F_{0,7}\left(h_0\right) \right)
\end{equation}
where the equivalent inverse equation computes
\begin{equation}
    h_{3} = h_{7} - \left(F_{2,7}\left(h_2\right) + F_{1,7}\left(h_1\right) + F_{0,7}\left(h_0\right) \right).
\end{equation}

\section{Cityscapes Semantic Segmentation}
\label{sec:exp:seg}

We present experimental results on the Cityscapes semantic segmentation dataset \cite{cordts2016cityscapes}.
The dataset contains 5000 high-quality pixel-level finely annotated images.
The finely annotated images are divided into 2,975/500/1,525 for training, validation, and testing.
There are 30 classes, and 19 among them are used for evaluation.
The mean of class-wise intersection over union (mIoU) is adopted as the evaluation metric.

\textbf{Setup.}
We follow the same training protocol as \citet{zhao2017pyramid,zhao2018psanet}.
The data are augmented by random cropping (from 1024 x 2048 to 512 x 1024), random scaling in the range [0.5, 2], and random horizontal flipping.
We use the SGD optimizer with a momentum of 0.9 and the weight decay of $5 * 10^{-4}$.
RevBiFPN-S0 through S3 use a base learning rate of 0.05; the rest of the networks use a base learning rate of 0.02.
The poly learning rate policy with a power of 0.9 is used with a minimum learning rate of $10^{-4}$.
All the models are trained for 90k iterations with a batch size of 16 across 8 GPUs and using SyncBN.
The models are trained using the MMSegmentation open-source segmentation toolbox \cite{mmseg2020}.
We use the provided FCN and OCR heads to evaluate the capability of the backbone for segmentation tasks.

\begin{table}
    \caption{
    Initial dropout and RandAugment ops applied (N).
    Training initially uses a weight decay of $4\times10^{-5}$, label smoothing = 0.1, a RandAugment magnitude of 9, and mstd is set to 0.5, and the network is trained without mixup, CutMix, or stochastic depth.
    }
    \vskip 10pt
    \centering
    \small
    \sc
    \begin{tabular}{lcc}
        \toprule
        Model        & DropOut &  N \\ \midrule
        RevBiFPN-S0 & 0.25    &  2 \\
        RevBiFPN-S1 & 0.25    &  2 \\
        RevBiFPN-S2 & 0.25    &  2 \\
        RevBiFPN-S3 & 0.25    &  2 \\
        RevBiFPN-S4 & 0.4     &  4 \\
        RevBiFPN-S5 & 0.4     &  4 \\
        RevBiFPN-S6 & 0.5     &  5 \\
        \bottomrule
    \end{tabular}
    \vskip -10pt
    \label{tab:regularization-i}
\end{table}

\textbf{Results.}
\cref{tab:cityscapes} shows the results of training RevBiFPN on the Cityscapes dataset in terms of parameters, computational complexity (evaluation MACs per sample\footnotemark[5]), memory used during training, and mIoU.
Although competitive, RevBiFPN can require larger model sizes to be as performant as the HRNet baseline.
Instead of using open-source training configurations and segmentation heads, future work can look at improving RevBiFPN's performance (per resource used) for semantic segmentation.

\begin{table*}
    \caption{
        Semantic segmentation results on the Cityscapes \texttt{val} set.
    }
    \vskip 10pt
    \centering
    \small
    \sc
    \begin{tabular}{l|rrrr|rrrr}
        \toprule
        \multirow{2}{*}{Backbone}        &  \multicolumn{4}{c|}{FCN Head}    & \multicolumn{4}{c}{ORC Head}    \\ 
                                         &  Params &     MACs &  Mem & mIoU &  Params &    MACs &  Mem & mIoU \\ \midrule
        RevBiFPN-S0                      &   1.88M &   52.72B & 0.65 & 72.8 &   5.66M &  503.5B & 2.71 & 73.9 \\
        RevBiFPN-S1                      &   2.97M &   85.41B & 0.74 & 74.1 &   7.49M &  632.6B & 2.95 & 75.7 \\
        RevBiFPN-S2                      &   7.00M &  200.78B & 1.60 & 75.6 &  13.35M &  990.5B & 3.46 & 76.9 \\
        RevBiFPN-S3                      &  14.36M &  354.64B & 2.01 & 77.4 &  22.34M & 1356.9B & 4.18 & 79.0 \\ 
        RevBiFPN-S4                      &  39.52M &  843.35B & 3.10 & 79.3 &  51.04M & 2309.5B & 5.47 & 80.4 \\ 
        RevBiFPN-S5                      &  68.04M & 1463.74B & 5.05 & 80.3 &  82.88M & 3364.7B & 6.21 & 80.8 \\
        RevBiFPN-S6                      & 122.56M & 2364.37B & 6.17 & 80.4 & 140.86M & 4719.6B & 6.44 & 81.8 \\ \midrule
        HRNetV2-W40~\cite{wang2020hrnet} &  45.89M &  536.46B & 3.31 & 80.2 &       - &       - &    - &    - \\
        HRNetV2-W48~\cite{wang2020hrnet} &  65.86M &  748.68B & 3.33 & 81.1 &   70.3M & 1206.3B & 4.84 & 81.6 \\
        \bottomrule
    \end{tabular}
    \label{tab:cityscapes}
\end{table*}


\begin{table*}
    \caption{
    Weight decay (WD), dropout, number of RandAugment ops applied (N), mixup,  CutMix, and stochastic depth used at the end of training.
    Label smoothing uses a coefficient of 0.1 and RandAugment uses a magnitude of 9 and mstd of 0.5.
    }
    \vskip 10pt
    \centering
    \small
    \sc
    \begin{tabular}{lcccccc}
        \toprule
        Model         &               WD & DropOut &  N & Mixup & CutMix & Stochastic Depth \\ \midrule
        RevBiFPN-S0  & $4\times10^{-5}$ & 0.25    &  2 &  0.00 &    0.0 &           0.00   \\
        RevBiFPN-S1  & $4\times10^{-5}$ & 0.25    &  2 &  0.00 &    0.0 &           0.00   \\
        RevBiFPN-S2  & $4\times10^{-5}$ & 0.3     &  3 &  0.00 &    0.0 &           0.00   \\
        RevBiFPN-S3  & $4\times10^{-5}$ & 0.3     &  3 &  0.10 &    1.0 &           0.05   \\
        RevBiFPN-S4  & $2\times10^{-5}$ & 0.4     &  4 &  0.10 &    1.0 &           0.10   \\
        RevBiFPN-S5  & $2\times10^{-5}$ & 0.4     &  4 &  0.20 &    1.0 &           0.10   \\
        RevBiFPN-S6  & $2\times10^{-5}$ & 0.6     &  5 &  0.20 &    1.0 &           0.30   \\
        \bottomrule
    \end{tabular}
    \label{tab:regularization-f}
\end{table*}

\section{ImageNet Regularization}
\label{appx:i1k_regularization}

Training is regularized using label smoothing~\cite{szegedy2016rethinking}, weight decay, dropout~\cite{srivastava2014dropout}, stochastic depth~\cite{huang2016deep}, CutMix~\cite{yun2019cutmix}, mixup~\cite{zhang2018mixup}, and the timm library \cite{wightman2019timm} variant of RandAugment~\cite{cubuk2020randaugment}.
Regularization increases with the network scale to prevent larger scales of the network from overfitting.
Without knowing how much augmentation was needed for each network, training began with the regularization in \cref{tab:regularization-i}.

When the validation accuracy of the EMA model began to plateau, the regularization of the models was increased. The final regularization used for each network is shown in \cref{tab:regularization-f}.

\section{ImageNet Model Comparisons}
\label{appx:i1k_comp}
\cref{tab:i1kacc_ext} extends \cref{tab:i1kacc}. This enable comparisons of RevBiFPN variants with other state  of the art networks trained on ImageNet-1K.

\begin{table*}
    \caption{Models trained using only ImageNet-1K. While most networks are trained using 300 to 400 epochs, HRNet and RegNetY use a 100 epoch training schedule.}
    \centering
    \small
    \sc
    \scalebox{0.78}{
    \begin{tabular}{lrrrrr}
        \toprule
        Model                                             & Params & Train Res & Res &  MACS &    Top1 \\ \midrule
        RevBiFPN-S0                                       &  3.42M &       224 & 224 & 0.31B & 72.8\% \\ 
        RevBiFPN-S1                                       &  5.11M &       256 & 256 & 0.62B & 75.9\% \\ 
        RevBiFPN-S2                                       &  10.6M &       256 & 256 & 1.37B & 79.0\% \\ 
        RevBiFPN-S3                                       &  19.6M &       288 & 288 & 3.33B & 81.1\% \\ 
        RevBiFPN-S4                                       &  48.7M &       320 & 320 & 10.6B & 83.0\% \\ 
        RevBiFPN-S5                                       &  82.0M &       352 & 352 & 21.8B & 83.7\% \\ 
        RevBiFPN-S6                                       & 142.3M &       352 & 352 & 38.1B & 84.2\% \\ 
        \midrule
        EfficientNet-B0~\cite{tan2019efficientnet}        &   5.3M &       224 & 224 & 0.39B & 77.1\%  \\
        EfficientNet-B1~\cite{tan2019efficientnet}        &   7.8M &       240 & 240 & 0.70B & 79.1\%  \\
        EfficientNet-B2~\cite{tan2019efficientnet}        &   9.2M &       260 & 260 &  1.0B & 80.1\%  \\
        EfficientNet-B3~\cite{tan2019efficientnet}        &    12M &       300 & 300 &  1.8B & 81.6\%  \\
        EfficientNet-B4~\cite{tan2019efficientnet}        &    19M &       380 & 380 &  4.2B & 82.9\%  \\
        EfficientNet-B5~\cite{tan2019efficientnet}        &    30M &       456 & 456 &  9.9B & 83.6\%  \\
        EfficientNet-B6~\cite{tan2019efficientnet}        &    43M &       528 & 528 &   19B & 84.0\%  \\
        EfficientNet-B7~\cite{tan2019efficientnet}        &    66M &       600 & 600 &   37B & 84.3\%  \\ \midrule
        EfficientNet-B5~\cite{cubuk2020randaugment}       &    30M &       456 & 456 &  9.9B & 83.9\%  \\
        EfficientNet-B7~\cite{cubuk2020randaugment}       &    66M &       600 & 600 &   37B & 85.0\%  \\ \midrule
        EfficientNetV2-S~\cite{tan2021efficientnetv2}     &    24M & 128 - 300 & 300 &  8.8B & 83.9\%  \\
        EfficientNetV2-M~\cite{tan2021efficientnetv2}     &    55M & 128 - 380 & 380 &   24B & 85.1\%  \\
        EfficientNetV2-L~\cite{tan2021efficientnetv2}     &   121M & 128 - 380 & 380 &   53B & 85.7\%  \\ \midrule
        NFNet-F0~\cite{brock2021high}                     &  72.0M &       192 & 256 &   12B & 83.6\%  \\
        NFNet-F1~\cite{brock2021high}                     &   133M &       224 & 320 &   36B & 84.7\%  \\
        NFNet-F2~\cite{brock2021high}                     &   194M &       256 & 352 &   63B & 85.1\%  \\
        NFNet-F3~\cite{brock2021high}                     &   255M &       320 & 416 &  115B & 85.7\%  \\
        NFNet-F4~\cite{brock2021high}                     &   316M &       384 & 512 &  215B & 85.9\%  \\ 
        NFNet-F5~\cite{brock2021high}                     &   377M &       416 & 544 &  290B & 86.0\%  \\ \midrule
        VOLO-D1~ \cite{yuan2021volo}                      &    27M &       224 & 384 & 22.8B & 85.2\%  \\
        VOLO-D2~ \cite{yuan2021volo}                      &    59M &       224 & 384 & 46.1B & 86.0\%  \\
        VOLO-D3~ \cite{yuan2021volo}                      &    86M &       224 & 448 & 67.9B & 86.3\%  \\
        VOLO-D4~ \cite{yuan2021volo}                      &   193M &       224 & 448 &  197B & 86.8\%  \\
        VOLO-D5~ \cite{yuan2021volo}                      &   269M &       224 & 448 &  304B & 87.0\%  \\
        VOLO-D5~ \cite{yuan2021volo}                      &   269M &       224 & 512 &  412B & 87.1\%  \\ \midrule
        ViT-B/16~\cite{dosovitskiy2020image}              &  86.0M &       384 & 384 & 55.4B & 77.91\% \\
        ViT-L/16~\cite{dosovitskiy2020image}              &   307M &       384 & 384 &  191B & 76.53\% \\ \midrule
        Swin-T~\cite{liu2021swin}                         &    29M &       224 & 224 &  4.5B & 81.3\%  \\
        Swin-S~\cite{liu2021swin}                         &    50M &       224 & 224 &  8.7B & 83.0\%  \\
        Swin-B~\cite{liu2021swin}                         &    88M &       224 & 384 & 47.0B & 84.5\%  \\ \midrule
        CoAtNet-0~\cite{dia2021coatnet}                   &    25M &       224 & 384 & 13.4B & 83.9\%  \\
        CoAtNet-1~\cite{dia2021coatnet}                   &    42M &       224 & 384 & 27.4B & 85.1\%  \\
        CoAtNet-2~\cite{dia2021coatnet}                   &    75M &       224 & 384 & 49.8B & 85.7\%  \\
        CoAtNet-2~\cite{dia2021coatnet}                   &    75M &       224 & 512 & 96.7B & 85.9\%  \\
        CoAtNet-3~\cite{dia2021coatnet}                   &   168M &       224 & 384 &  107B & 85.8\%  \\
        CoAtNet-3~\cite{dia2021coatnet}                   &   168M &       224 & 512 &  203B & 86.0\%  \\ \midrule
        CaiT-XXS-24~\cite{touvron2021going}               &  12.0M &       224 & 384 &  9.5B & 80.4\%  \\
        CaiT-XXS-36~\cite{touvron2021going}               &  17.3M &       224 & 384 & 14.2B & 81.8\%  \\
        CaiT-XS-24~\cite{touvron2021going}                &  26.6M &       224 & 384 & 19.3B & 83.8\%  \\
        CaiT-XS-36~\cite{touvron2021going}                &  38.6M &       224 & 384 & 28.8B & 84.3\%  \\
        CaiT-S-24~\cite{touvron2021going}                 &  46.9M &       224 & 384 & 32.2B & 84.3\%  \\
        CaiT-S-36~\cite{touvron2021going}                 &  68.2M &       224 & 384 & 48.0B & 85.0\%  \\
        CaiT-S-48~\cite{touvron2021going}                 &  89.5M &       224 & 384 & 63.8B & 85.1\%  \\
        CaiT-M-24~\cite{touvron2021going}                 & 185.9M &       224 & 384 &116.1B & 84.5\%  \\
        CaiT-M-36~\cite{touvron2021going}                 & 270.9M &       224 & 384 &173.3B & 84.9\%  \\ \midrule
        HRNet-W18-C~\cite{sun2019hrnet_pose}              &  21.3M &       224 & 224 & 3.99B & 76.8\%  \\
        HRNet-W30-C~\cite{sun2019hrnet_pose}              &  37.7M &       224 & 224 & 7.55B & 78.2\%  \\
        HRNet-W32-C~\cite{sun2019hrnet_pose}              &  41.2M &       224 & 224 & 8.31B & 78.5\%  \\
        HRNet-W40-C~\cite{sun2019hrnet_pose}              &  57.6M &       224 & 224 & 11.8B & 78.9\%  \\
        HRNet-W44-C~\cite{sun2019hrnet_pose}              &  67.1M &       224 & 224 & 13.9B & 78.9\%  \\
        HRNet-W48-C~\cite{sun2019hrnet_pose}              &  77.5M &       224 & 224 & 16.1B & 79.3\%  \\
        HRNet-W64-C~\cite{sun2019hrnet_pose}              &   128M &       224 & 224 & 26.9B & 79.5\%  \\ \midrule
        RegNetY-200MF~\cite{radosavovic2020designing}     &   3.2M &       224 & 224 &  0.2B & 70.4\%  \\
        RegNetY-400MF~\cite{radosavovic2020designing}     &   4.3M &       224 & 224 &  0.4B & 74.1\%  \\
        RegNetY-600MF~\cite{radosavovic2020designing}     &   6.1M &       224 & 224 &  0.6B & 75.5\%  \\
        RegNetY-800MF~\cite{radosavovic2020designing}     &   6.3M &       224 & 224 &  0.8B & 76.3\%  \\
        RegNetY-1.6GF~\cite{radosavovic2020designing}     &  11.2M &       224 & 224 &  1.6B & 78.0\%  \\
        RegNetY-3.2GF~\cite{radosavovic2020designing}     &  19.4M &       224 & 224 &  3.2B & 79.0\%  \\
        RegNetY-4.0GF~\cite{radosavovic2020designing}     &  20.6M &       224 & 224 &  4.0B & 79.4\%  \\
        RegNetY-6.4GF~\cite{radosavovic2020designing}     &  30.6M &       224 & 224 &  6.4B & 79.9\%  \\
        RegNetY-8.0GF~\cite{radosavovic2020designing}     &  39.2M &       224 & 224 &  8.0B & 79.9\%  \\
        RegNetY-12GF~\cite{radosavovic2020designing}      &  51.8M &       224 & 224 & 12.1B & 80.3\%  \\
        RegNetY-16GF~\cite{radosavovic2020designing}      &  83.6M &       224 & 224 & 15.9B & 80.4\%  \\
        RegNetY-32GF~\cite{radosavovic2020designing}      & 145.0M &       224 & 224 & 32.3B & 81.0\%  \\
        \bottomrule
    \end{tabular}
    }
    \label{tab:i1kacc_ext}
\end{table*}

\section{MS COCO Extended Results}
\label{sec:exp:mscoco_ext_results}

\cref{tab:cocofrcnn,tab:cocomrcnn} provide the numerical details of \cref{fig:memvsap_faster,fig:memvsap_mask,fig:macvsap_faster,fig:macvsap_mask,fig:paramvsap_faster,fig:paramvsap_mask} extending the results of \cref{sec:exp:coco}.

\begin{table*}
    \caption{
        Object detection results on COCO
        \texttt{minival} 
        in the Faster R-CNN framework.
        LS = learning schedule.
        1x = 12 epochs, 2x = 24 epochs.
        Mem = GPU memory used during training.
        RevBiFPN performs better than HRNet and ResNet
        on small (AP\textsubscript{S}), medium (AP\textsubscript{M}), and large (AP\textsubscript{L}) objects
        while using fewer MACs and less training-time memory.
        Evaluation MACs per sample are reported using the training image resolution (800$\times$1333).
    }
    \vskip 10pt
    \centering
    \small
    \sc
    \resizebox{\textwidth}{!}{
    \begin{tabular}{lrrrrrrrrrr}
        \toprule
        Backbone                            \hspace{-10pt} &  Params &    MACs &      Mem & \hspace{-6pt} LS \hspace{-5pt} & AP   & AP$_{50}$ & AP$_{75}$ & AP$_{S}$ & AP$_{M}$ & AP$_{L}$ \\ \midrule
        RevBiFPN-S0                         \hspace{-10pt} &  19.55M & 135.12B &  1.67 GB & \hspace{-6pt} 1x \hspace{-5pt} & 31.4 &      51.5 &      33.3 &     17.8 &     34.3 &     40.9 \\
        RevBiFPN-S1                         \hspace{-10pt} &  20.48M & 140.66B &  1.78 GB & \hspace{-6pt} 1x \hspace{-5pt} & 32.0 &      52.0 &      34.1 &     18.3 &     35.7 &     43.0 \\
        RevBiFPN-S2                         \hspace{-10pt} &  23.86M & 157.42B &  2.13 GB & \hspace{-6pt} 1x \hspace{-5pt} & 36.3 &      57.4 &      39.3 &     20.8 &     39.6 &     46.6 \\
        RevBiFPN-S3                         \hspace{-10pt} &  30.40M & 180.99B &  2.61 GB & \hspace{-6pt} 1x \hspace{-5pt} & 38.7 &      60.0 &      41.4 &     23.1 &     42.0 &     50.4 \\
        RevBiFPN-S4                         \hspace{-10pt} &  52.88M & 251.02B &  4.05 GB & \hspace{-6pt} 1x \hspace{-5pt} & 40.3 &      60.5 &      44.0 &     23.7 &     44.3 &     52.4 \\ 
        RevBiFPN-S5                         \hspace{-10pt} &  77.83M & 328.91B &  5.50 GB & \hspace{-6pt} 1x \hspace{-5pt} & 41.3 &      62.7 &      44.8 &     24.8 &     45.6 &     52.5 \\ 
        RevBiFPN-S6                         \hspace{-10pt} & 127.51M & 465.43B &  7.37 GB & \hspace{-6pt} 1x \hspace{-5pt} & 42.2 &      63.5 &      45.8 &     25.7 &     46.5 &     54.0 \\ \midrule
        HRNetV2p-W18~\cite{wang2020hrnet}   \hspace{-10pt} &  27.48M & 196.18B &  6.25 GB & \hspace{-6pt} 1x \hspace{-5pt} & 36.2 &      57.3 &      39.3 &     20.7 &     39.0 &     46.8 \\
        HRNetV2p-W18~\cite{wang2020hrnet}   \hspace{-10pt} &  27.48M & 196.18B &  6.25 GB & \hspace{-6pt} 2x \hspace{-5pt} & 38.0 &      58.9 &      41.5 &     22.6 &     40.8 &     49.6 \\
        HRNetV2p-W32~\cite{wang2020hrnet}   \hspace{-10pt} &  47.28M & 298.96B &  8.62 GB & \hspace{-6pt} 1x \hspace{-5pt} & 39.6 &      61.0 &      43.3 &     23.7 &     42.5 &     50.5 \\ 
        HRNetV2p-W32~\cite{wang2020hrnet}   \hspace{-10pt} &  47.28M & 298.96B &  8.62 GB & \hspace{-6pt} 2x \hspace{-5pt} & 40.9 &      61.8 &      44.8 &     24.4 &     43.7 &     53.3 \\ 
        HRNetV2p-W48~\cite{wang2020hrnet}   \hspace{-10pt} &  83.36M & 481.92B & 11.64 GB & \hspace{-6pt} 1x \hspace{-5pt} & 41.3 &      62.8 &      45.1 &     25.1 &     44.5 &     52.9 \\ 
        HRNetV2p-W48~\cite{wang2020hrnet}   \hspace{-10pt} &  83.36M & 481.92B & 11.64 GB & \hspace{-6pt} 2x \hspace{-5pt} & 41.8 &      62.8 &      45.9 &     25.0 &     44.7 &     54.6 \\ \midrule
        ResNet-50-FPN~\cite{wang2020hrnet}  \hspace{-10pt} &  41.53M & 216.70B &  3.61 GB & \hspace{-6pt} 1x \hspace{-5pt} & 36.7 &      58.3 &      39.9 &     20.9 &     39.8 &     47.9 \\
        ResNet-50-FPN~\cite{wang2020hrnet}  \hspace{-10pt} &  41.53M & 216.70B &  3.61 GB & \hspace{-6pt} 2x \hspace{-5pt} & 37.6 &      58.7 &      41.3 &     21.4 &     40.8 &     49.7 \\
        ResNet-101-FPN~\cite{wang2020hrnet} \hspace{-10pt} &  60.52M & 296.58B &  5.43 GB & \hspace{-6pt} 1x \hspace{-5pt} & 39.2 &      61.1 &      43.0 &     22.3 &     42.9 &     50.9 \\
        ResNet-101-FPN~\cite{wang2020hrnet} \hspace{-10pt} &  60.52M & 296.58B &  5.43 GB & \hspace{-6pt} 2x \hspace{-5pt} & 39.8 &      61.4 &      43.4 &     22.9 &     43.6 &     52.4 \\
        \bottomrule
    \end{tabular}
    }
    \label{tab:cocofrcnn}
\end{table*}

\begin{table*}
    \caption{
        Instance segmentation and detection results on COCO 
        \texttt{minival} 
        in the Mask R-CNN framework. 
        LS = learning schedule.
        1x = 12 epochs, 2x = 24 epochs.
        Mem = GPU memory used during training.
        RevBiFPN outperforms HRNet
        for detection and segmentation 
        small (AP\textsubscript{S}) and medium (AP\textsubscript{M}) objects, as well as bounding box AP for medium (AP\textsubscript{M}) and large (AP\textsubscript{L}) objects, 
        while using fewer MACs and less training-time memory. 
        Evaluation MACs per sample are reported using the training image resolution (800$\times$1333).
    }
    \vskip 10pt
    \centering
    \small
    \sc
    \resizebox{\textwidth}{!}{
    \begin{tabular}{lrrrrrrrrrrrr}
        \toprule
        \multirow{2}{*}{Backbone}           \hspace{-10pt} & \multirow{2}{*}{Params} & \multirow{2}{*}{MACs} & \multirow{2}{*}{Mem} & \hspace{-6pt} \multirow{2}{*}{LS} \hspace{-5pt} & \multicolumn{4}{c}{Mask}              & \multicolumn{4}{c}{BBox}              \\ 
                                            \hspace{-10pt} &                         &                       &                      & \hspace{-6pt}                     \hspace{-5pt} &   AP & AP$_{S}$ & AP$_{M}$ & AP$_{L}$ &   AP & AP$_{S}$ & AP$_{M}$ & AP$_{L}$ \\ \midrule
        RevBiFPN-S0                         \hspace{-10pt} &                   22.2M &                188.2B &                2.0GB & \hspace{-6pt}                  1x \hspace{-5pt} & 29.7 &     13.5 &     32.3 &     44.2 & 31.4 &     17.8 &     34.3 &     40.9 \\
        RevBiFPN-S1                         \hspace{-10pt} &                   23.1M &                193.7B &                2.4GB & \hspace{-6pt}                  1x \hspace{-5pt} & 31.0 &     14.1 &     33.3 &     45.3 & 34.0 &     19.0 &     37.0 &     44.6 \\
        RevBiFPN-S2                         \hspace{-10pt} &                   26.5M &                210.5B &                2.6GB & \hspace{-6pt}                  1x \hspace{-5pt} & 33.7 &     16.0 &     35.9 &     49.2 & 37.1 &     21.7 &     40.2 &     48.5 \\
        RevBiFPN-S3                         \hspace{-10pt} &                   33.0M &                232.9B &                2.6GB & \hspace{-6pt}                  1x \hspace{-5pt} & 35.5 &     17.4 &     38.4 &     50.9 & 39.4 &     23.6 &     43.1 &     50.9 \\ 
        RevBiFPN-S4                         \hspace{-10pt} &                   55.5M &                304.1B &                4.1GB & \hspace{-6pt}                  1x \hspace{-5pt} & 37.1 &     17.8 &     40.1 &     53.4 & 41.5 &     24.2 &     45.4 &     53.9 \\ 
        RevBiFPN-S5                         \hspace{-10pt} &                   80.5M &                382.0B &                5.5GB & \hspace{-6pt}                  1x \hspace{-5pt} & 37.8 &     18.5 &     40.7 &     54.3 & 42.2 &     25.5 &     46.3 &     54.3 \\
        RevBiFPN-S6                         \hspace{-10pt} &                  130.2M &                518.5B &                7.4GB & \hspace{-6pt}                  1x \hspace{-5pt} & 38.7 &     19.8 &     41.7 &     55.2 & 43.3 &     26.9 &     47.4 &     55.6 \\ \midrule
        HRNetV2p-W18~\cite{wang2020hrnet}   \hspace{-10pt} &                   30.1M &                249.3B &                6.7GB & \hspace{-6pt}                  1x \hspace{-5pt} & 33.8 &     15.6 &     35.6 &     49.8 & 37.1 &     21.9 &     39.5 &     47.9 \\
        HRNetV2p-W18~\cite{wang2020hrnet}   \hspace{-10pt} &                   30.1M &                249.3B &                6.7GB & \hspace{-6pt}                  2x \hspace{-5pt} & 35.3 &     16.9 &     37.5 &     51.8 & 39.2 &     23.7 &     41.7 &     51.0 \\
        HRNetV2p-W32~\cite{wang2020hrnet}   \hspace{-10pt} &                   49.9M &                352.0B &                9.0GB & \hspace{-6pt}                  1x \hspace{-5pt} & 36.7 &     17.3 &     39.0 &     53.0 & 40.9 &     24.5 &     43.9 &     52.2 \\
        HRNetV2p-W32~\cite{wang2020hrnet}   \hspace{-10pt} &                   49.9M &                352.0B &                9.0GB & \hspace{-6pt}                  2x \hspace{-5pt} & 37.6 &     17.8 &     40.0 &     55.0 & 42.3 &     25.0 &     45.4 &     54.9 \\ \midrule
        ResNet-50-FPN~\cite{wang2020hrnet}  \hspace{-10pt} &                   44.2M &                269.8B &                4.2GB & \hspace{-6pt}                  1x \hspace{-5pt} & 34.2 &     15.7 &     36.8 &     50.2 & 37.8 &     22.1 &     40.9 &     49.3 \\
        ResNet-50-FPN~\cite{wang2020hrnet}  \hspace{-10pt} &                   44.2M &                269.8B &                4.2GB & \hspace{-6pt}                  2x \hspace{-5pt} & 35.0 &     16.0 &     37.5 &     52.0 & 38.6 &     21.7 &     41.6 &     50.9 \\
        ResNet-101-FPN~\cite{wang2020hrnet} \hspace{-10pt} &                   63.2M &                349.7B &                5.8GB & \hspace{-6pt}                  1x \hspace{-5pt} & 36.1 &     16.2 &     39.0 &     53.0 & 40.0 &     22.6 &     43.4 &     52.3 \\
        ResNet-101-FPN~\cite{wang2020hrnet} \hspace{-10pt} &                   63.2M &                349.7B &                5.8GB & \hspace{-6pt}                  2x \hspace{-5pt} & 36.7 &     17.0 &     39.5 &     54.8 & 41.0 &     23.4 &     44.4 &     53.9 \\
        \bottomrule
    \end{tabular}
    }
    \label{tab:cocomrcnn}
\end{table*}


\end{document}